\documentclass[lettersize,journal]{IEEEtran}
\usepackage{amsmath,amsfonts}
\usepackage{algorithmic}
\usepackage{algorithm}
\usepackage{array}
\usepackage[caption=false,font=normalsize,labelfont=sf,textfont=sf]{subfig}
\usepackage{textcomp}
\usepackage{stfloats}
\usepackage{url}
\usepackage{verbatim}
\usepackage{graphicx}
\usepackage{diagbox}
\usepackage{booktabs}
\usepackage{multirow}
\usepackage{cite}
\begin{document}

\title{Robot Deformable Object Manipulation via NMPC-generated Demonstrations in Deep Reinforcement Learning}

\author{Haoyuan Wang,
	Zihao Dong,
	Hongliang Lei,
	Zejia Zhang,
	Weizhuang Shi,
	Wei Luo, 
	Weiwei Wan, ~\IEEEmembership{Senior Member,~IEEE,}
	and Jian Huang ~\IEEEmembership{Senior Member,~IEEE}
	\thanks{This work was supported in part by National Natural Science Foundation of China under Grant 62333007 and U24A20280, and in part by Hubei Provincial Technology Innovation Program under Grant 2024BAA007. H. Wang and Z. Dong contributed equally to this work. \textit{(Corresponding authors: Jian Huang; Wei Luo.)} }
	\thanks{H. Wang, Z. Dong, H. Lei, Z. Zhang, W. Shi, and J. Huang are with Hubei Key Laboratory of Brain-inspired Intelligent Systems and the Key Laboratory of Image Processing and Intelligent Control, School of Artificial Intelligence and Automation, Huazhong University of Science and	Technology (HUST), Wuhan 430074, Hubei, China. Email: {\tt\small \{why427@, M202173202@,  leihl@, zejiazhang@, swz@, huang\_jan@mail.\}hust.edu.cn}}
	\thanks{W. Luo is with Department of Innovation Center, China Ship Development and Design Center, Wuhan 430064, Hubei, China. Email: {\tt\small csddc\_weiluo@mail.163.com}}
	\thanks{W. Wan is with the Graduate School of Engineering Science, Osaka University, Toyonaka 560-0043, Japan. Email: {\tt\small wan@sys.es.osaka-u.ac.jp}}
}



\maketitle

\begin{abstract}
In this work, we conducted research on deformable object manipulation by robots based on demonstration-enhanced reinforcement learning (RL). To improve the learning efficiency of RL, we enhanced the utilization of demonstration data from multiple aspects and proposed the HGCR-DDPG algorithm. It uses a novel high-dimensional fuzzy approach for grasping-point selection, a refined behavior-cloning method to enhance data-driven learning in Rainbow-DDPG, and a sequential policy-learning strategy. Compared to the baseline algorithm (Rainbow-DDPG), our proposed HGCR-DDPG achieved 2.01 times the global average reward and reduced the global average standard deviation to 45\% of that of the baseline algorithm. To reduce the human labor cost of demonstration collection, we proposed a low-cost demonstration collection method based on Nonlinear Model Predictive Control (NMPC).
Simulation experiment results show that demonstrations collected through NMPC can be used to train HGCR-DDPG, achieving comparable results to those obtained with human demonstrations. To validate the feasibility of our proposed methods in real-world environments, we conducted physical experiments involving deformable object manipulation. We manipulated fabric to perform three tasks: diagonal folding, central axis folding, and flattening. The experimental results demonstrate that our proposed method achieved success rates of 83.3\%, 80\%, and 100\% for these three tasks, respectively, validating the effectiveness of our approach. Compared to current large-model approaches for robot manipulation, the proposed algorithm is lightweight, requires fewer computational resources, and offers task-specific customization and efficient adaptability for specific tasks.
\end{abstract}

\begin{IEEEkeywords}
Deformable objects, robotic Manipulation, reinforcement Learning, demonstration, nonlinear model predictive control
\end{IEEEkeywords}

\section{Introduction}
\IEEEPARstart{D}{eformable} objects play a critical role in numerous key industries and are widely used across various sectors \cite{zhu1}. Their manipulation is a common task in manufacturing \cite{c1,c2}, medical surgery \cite{c3,c4}, and service robotics \cite{c5,c6,c7}. However, manual handling of deformable objects can be time-consuming, labor-intensive, costly, and may not guarantee efficiency or accuracy. Consequently, robots are often employed to replace human operators for manipulating deformable objects, such as connecting cables on automated assembly lines \cite{c8, zhu3}, cutting or suturing soft tissue during medical surgeries \cite{c9}, and handling fabrics like towels and clothes in home service scenarios \cite{c10, zhu2}. Automating the manipulation of deformable objects with robots can significantly reduce labor costs while improving operational efficiency and precision. Therefore, robotic systems for manipulating deformable objects have attracted considerable attention and research \cite{c11}.

Currently, the majority of robotic manipulation research focuses on rigid objects, where the deformation caused during grasping is negligible. In contrast, when dealing with deformable objects, robots face many new challenges, including high-dimensional state spaces, complex dynamics, and highly nonlinear physical properties \cite{c12}. To address these challenges, some researchers have established dynamical models for deformable objects and designed robotic manipulation strategies based on these models \cite{c13, c14}.  Nonetheless, ensuring high accuracy in the dynamical model presents significant difficulties, and the derivation of model gradients can be highly complex \cite{c12,c15}. To avoid the complexity of dynamical model derivation, some researchers have turned to learning-based methods, particularly reinforcement learning (RL) and imitation learning (IL) \cite{c12}. These methods learn control policies from data using learning algorithms, without requiring explicit dynamical modeling.
RL involves the agent continuously exploring the action space through trial and error, collecting interaction data with the environment to facilitate learning. Still, in real-world scenarios, the intricacy involved in handling deformable objects frequently results in inefficient learning processes that require extensive time and data, yielding less-than-optimal results. Therefore, it is crucial to set more effective states and action spaces to reduce task complexity based on domain-specific knowledge \cite{c16,c17}. With the advancement of deep learning technology, Deep Reinforcement Learning (DRL) is being used to tackle deformable object manipulation problems. Matas et al. \cite{c18} trained agents using DRL methods in simulation environments to fold clothes or place them on hangers. Researchers incorporated seven commonly used techniques in the DRL field into the Deep Deterministic Policy Gradient (DDPG) to develop the Rainbow-DDPG algorithm and validated the effectiveness of these techniques through ablation experiments. Additionally, they conducted deformable object manipulation experiments in real scenes through domain randomization. Jangir et al. \cite{c19} treated the coordinates of a series of key points on the fabric as the state space, introducing Behavioral Cloning (BC) and Hindsight Experience Replay (HER) to improve the DDPG algorithm for handling tasks involving dynamic manipulation of fabrics by robots. They also studied the impact of key point selection on RL performance. 
Despite making some progress in learning effective strategies for deformable object manipulation, DRL still faces challenges in terms of learning efficiency due to the inherent complexity of such manipulation tasks, requiring substantial amounts of data and computational resources for training. 

Collecting human demonstration data and extracting operational skills using IL algorithms from these demonstrations has also received extensive research attention. Unlike the trial-and-error mechanism of RL, IL completes tasks through observation and imitation of expert behavior. This method has unique advantages in handling tasks that are too complex for RL or where clear rewards are difficult to define \cite{c23}. 
With the continuous development of deep learning technology and hardware infrastructure, recent research has been able to collect a large amount of human demonstration data and utilize deep learning techniques to extract manipulation skills from it \cite{c24,c25, zhu4}. Although extracting manipulation skills from a large amount of human demonstrations can yield decent results, the high manpower cost associated with this approach is often challenging and unsustainable. Some studies combine RL and IL, leveraging human demonstrations to enhance the learning efficiency of RL while also benefiting from RL's ability for autonomous exploration \cite{c18,c19,c26}. Balaguer et al. \cite{c26} used the K-means clustering algorithm to categorize human demonstration actions into $\mathit{M}$ classes, testing the feasibility of each type of human demonstration action on the robot and selecting the feasible one with the highest reward as the starting point for agent exploration, thus streamlining the search space of RL algorithms. It is worth mentioning that the Rainbow-DDPG algorithm mentioned earlier \cite{c18} and the work by Jangir et al. \cite{c19} also incorporate human demonstration data to improve the learning efficiency of RL. Undeniably, the morphological diversity of deformable objects imposes higher requirements on the range of operational scenarios covered by demonstration data. To cover as wide a range of operational scenarios as possible, researchers typically need to collect a large amount of demonstration data. Existing studies \cite{c24,c25} use manually collected large-scale demonstration data for training purposes, which inevitably incurs significant human resource costs.

To address the challenges of low learning efficiency in RL and the difficulty in collecting IL demonstration data mentioned above, this article adopts a learning-based approach to tackle the problem of deformable object manipulation by robots. The aim is to improve the learning efficiency of algorithms and reduce learning costs. The research aims to optimize existing RL algorithms from two perspectives. First, by integrating IL, we innovatively design the HGCR-DDPG algorithm. It leverages a novel high-dimensional fuzzy-based approach to select grasping points, a refined behavior cloning-inspired method to boost data-driven learning in Rainbow-DDPG, and a sequential policy-learning strategy. This holistic design enhances RL learning efficiency. Second, we develop a low-cost demonstration data collection method using NMPC. It's built upon a spring-mass model, enabling automated data generation and effective mapping to robot actions, thus reducing data collection costs. Through these, robots can learn manipulation skills with higher efficiency and lower cost, thus operating deformable objects more efficiently in practical applications. Specifically, the main contributions of this article are as follows:

\begin{itemize}
	\item{An RL method enhanced by demonstration to increase the learning efficiency of RL with human demonstration data for training, which is named as HGCR-DDPG.}
	\item{A demonstration data collection method in simulation environment based on Nonlinear Model Predictive Control (NMPC) to reduce the cost of demonstration data collection.}
	\item{The feasibility of the proposed methods in simulation and real environments through experiments.}
\end{itemize}

\begin{figure*}[tb]
	\centering
	\includegraphics[width=0.77\textwidth]{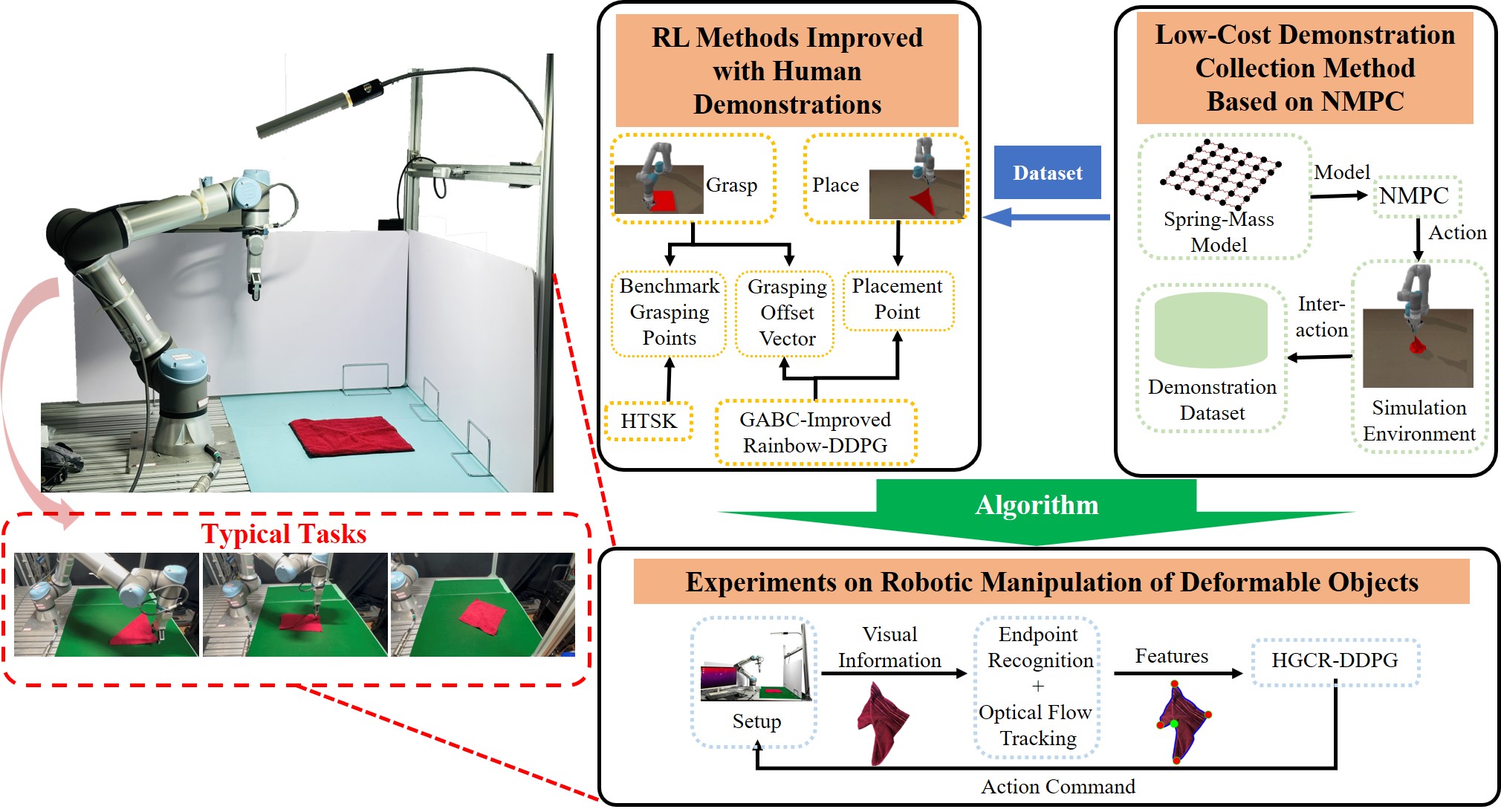}
	\caption{The diagram of this work.}
	\label{The diagram of this work}
\end{figure*}

The article is divided into 6 sections, with the main research content and their relationships shown in Fig. \ref{The diagram of this work}. The specific content arrangement of each section is as follows: Section I is the introduction. Section II addresses the issue of low learning efficiency in RL algorithms by proposing HGCR-DDPG algorithm that combines a High-Dimensional Takagi-Sugeno-Kang (HTSK) fuzzy system, Generative Adversarial Behavior Cloning (GABC) techniques, Rainbow-DDPG, and Conditional Policy Learning (CPL). Section III addresses the issue of high cost of collecting demonstration data by exploring automated demonstration collection techniques and proposes a low-cost demonstration collection method based on NMPC. Section IV presents the simulation and physical experiment settings that validate the methods proposed in this article. Section V presents the experiment results. Section VI provides a summary of the entire article and outlooks future work.

\section{Problem Formulation}
\subsection{Introduction to the Simulation Environment and the Physical Experiment Platform}


A complex deformable object manipulation simulation environment was established using PyBullet \cite{c30}. Specifically, a UR5e robot model and a square cloth model with a side length of 0.24 meters were constructed. The cloth model consisted of detailed triangular meshes, with its mass uniformly distributed across the mesh vertices (i.e., nodes). The process of the robot grasping the cloth was simulated by establishing precise anchor connections between the corresponding nodes of the cloth model and the robot end-effector. During the simulation, the cloth model was affected only by gravity, friction with the table surface, and the traction force exerted by the robot. The simulation experiments were conducted on a high-performance server equipped with 48GB of RAM, an Intel E5-2660 v4 processor, and an NVIDIA 3090 graphics card.

The deformable object manipulation system in this article consists of sensor subsystems, decision and control subsystems, and robot motion subsystems, as illustrated in Fig. S1(a). The target object for the experiment is a red square fabric with a side length of 0.24 meters. The sensor subsystem is responsible for capturing image information of the fabric.
The robot motion subsystem executes precise grasping and placing actions. The decision and control subsystem extract key features from visual information, utilize the algorithm to generate decision commands, and communicate with the robot motion subsystem via ROS.
As depicted in Fig. S1(b), the Intel RealSense D435i camera is fixed in an ``eye-to-hand" manner, 
the UR5e robotic arm is mounted on a dedicated workstation, 
and the RG2 gripper, serving as the execution end, is installed at the end of the robot.

%

\subsection{Problem Formulation for Robotic Manipulation of Deformable Objects under the DRL Framework}

This section will provide a detailed introduction to the DRL model of this study from five aspects: task setting, state space, action space, state transition, and reward setting.

\subsubsection{Task Setting}
For a square piece of fabric as a representative deformable object, these tasks are designed: folding along the diagonal, folding along the central axis, and flattening. In each task, the robot is allowed a maximum number of operations, denoted as $t_\text{m}$, which varies among different tasks. At the beginning of each experimental round, the robot returns to its default initial pose, while the initial position of the fabric is set according to the requirements of the specific task and is subject to a certain degree of random noise. 
The details are as followed:

\begin{itemize}
	\item{\textbf{Folding along the diagonal:} 
		The specific objective of the operation is to achieve perfect alignment of one pair of diagonal endpoints of the fabric, while maintaining the distance between the other pair of diagonal endpoints exactly equal to the length of the fabric diagonal, and ensuring that the area of the fabric is equal to half of its area when fully unfolded.}
	\item{\textbf{Folding along the central axis:} 
		Before folding, the two sets of fabric endpoints should be symmetrically arranged relative to the folding axis. After folding, these two sets of endpoints should coincide, while ensuring that the distance between endpoints on the side of the folding axis remains consistent with before folding, and the area of the fabric is equal to half of its area when fully unfolded.}
	\item{\textbf{Flattening:} 
		When faced with heavily wrinkled fabric, the robot's task is to flatten it to its maximum area. At the beginning of each experiment, the fabric is initialized by the robot applying random actions within the first 10 time steps. The robot moves a point on the fabric from its initial position to a placement point within a distance of 0.1 m to 0.2 m during each random step, ensuring sufficient disturbance to generate random wrinkles. 
	}
\end{itemize}
%

\subsubsection{State Space}
Previous studies \cite{c18,c27} often directly fed the visual information of the scene as state inputs to DRL, which is intuitive but results in an overly complex state space. Some research \cite{c19} simplifies the state space in simulation by using the coordinates of deformable object feature points as state inputs, which is simple but difficult to directly transfer to the real world. 
This article adopts a compromise solution. Algorithms from OpenCV are utilized to preprocess the visual information of the scene, and the processed results are then used as state inputs for DRL. The state spaces of three different tasks are introduced as follows.

\begin{figure}[tb]
	\centering
	\includegraphics[width=0.45\textwidth]{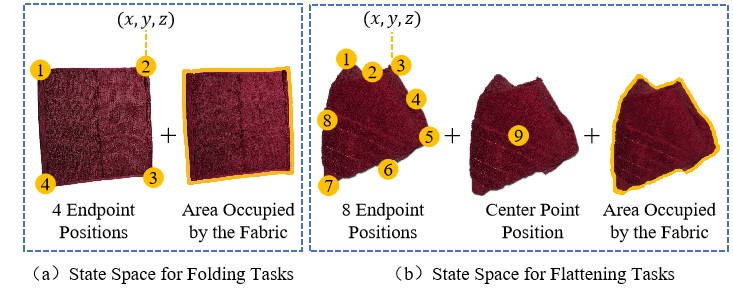}
	\caption{State Spaces for Different Tasks.}
	\label{State Spaces for Different Tasks}
\end{figure}

In both the along-diagonal and along-axis folding tasks, 
using Canny edge detection \cite{c31} and the Douglas-Peucker algorithm in OpenCV, the four right-angle corners of the fabric can be identified. During robot manipulation, considering the relationship between the fabric's motion speed and the robot's operation speed, this article, under the premise of relatively slow robot operation, employs the pyramid Lucas-Kanade optical flow tracking method to track the four corners. 
This article selects the positions of the four corners of the square fabric and the proportion $f_t$ of the fabric's current area to its area when fully flattened as the state representation in these two folding tasks.
This results in a 13-dimensional state space, as shown in Fig. \ref{State Spaces for Different Tasks}(a). The symbols defining the state variables for the folding tasks are as follows:

\begin{equation}
	\boldsymbol{s}_t=(\boldsymbol{p}_{1_t},\boldsymbol{p}_{2_t},\boldsymbol{p}_{3_t},\boldsymbol{p}_{4_t},f_t),
	\label{eq:fold-state}
\end{equation}
where $\boldsymbol{p}_{1_t},\boldsymbol{p}_{2_t},\boldsymbol{p}_{3_t},\boldsymbol{p}_{4_t}$ respectively represent the three-dimensional coordinates of the four corners of the fabric at time $t$. All coordinates and vectors mentioned in this article are described with respect to the base coordinate system of the robot they are associated with.

In the flattening task, the fabric's initial state is heavily wrinkled, making it extremely difficult to detect the right-angle corners of the fabric. 
We use Canny edge detection and the Douglas-Peucker algorithm in OpenCV to fit the contour of the fabric into an octagon, representing the eight points on the contour that best characterize the shape of the fabric. 
The coordinates of the eight endpoints, the coordinates of the center point of the fabric contour and the proportion of the fabric's current area to its area when fully flattened are the state representation.
Ultimately, the dimensionality of the state space used in the spreading task in this article is 28, as shown in Fig. \ref{State Spaces for Different Tasks}(b). The symbols defining the state variables for the spreading task are as follows:

\begin{equation}
	\boldsymbol{s}_t=(\boldsymbol{p}_{1_t},\boldsymbol{p}_{2_t},\cdots,\boldsymbol{p}_{8_t},\boldsymbol{p}_{\text{c}_t},f_t),
	\label{eq:flatten-state}
\end{equation}
where $\boldsymbol{p}_{1_t},\boldsymbol{p}_{2_t},\cdots,\boldsymbol{p}_{8_t}$ respectively represent the three-dimensional coordinates of the eight fitted endpoints of the fabric at time $t$, and $\boldsymbol{p}_{\text{c}_t}$ represents the three-dimensional coordinates of the center point of the fabric contour at time $t$.

\subsubsection{Action Space}
During the process of collecting human demonstrations, we found that humans only need to manipulate the four endpoints of the fabric to complete all folding tasks. 
In contrast, a strategy solely based on manipulating the endpoints of the fabric outline has minimal effect in flattening tasks. Fig. S1 explains this phenomenon: the fabric in a folded state is divided into upper layer (green), intermediate connecting parts (purple), and lower layer (orange), as shown in Fig. S2(a). Effective relative displacement between the upper and lower layers occurs only when manipulating points on the upper layer, as depicted in Fig. S2(b). Conversely, manipulating the lower layer or the connecting parts, as shown in Fig. S2(c), mostly results in overall movement of the fabric, which is not substantially helpful for flattening tasks.

\begin{figure}[tb]
	\centering
	\includegraphics[width=0.35\textwidth]{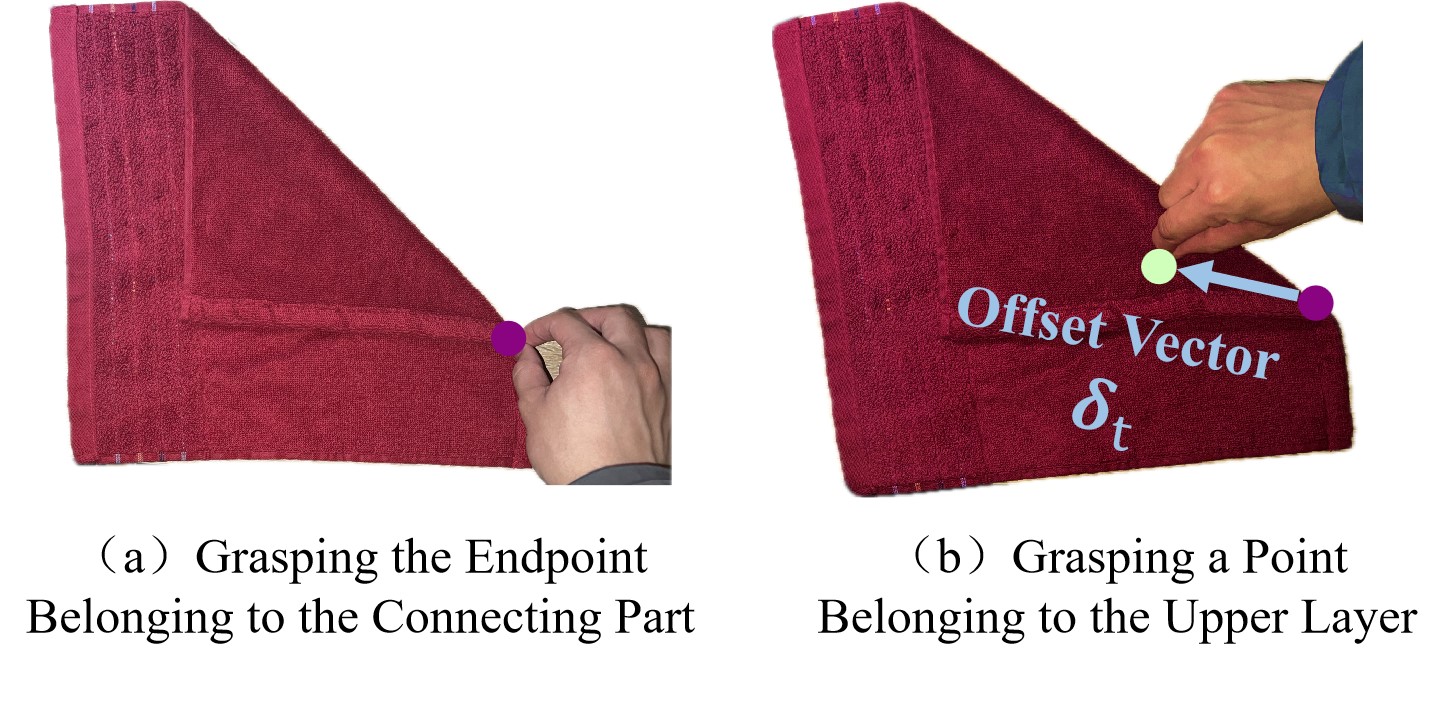}
	\caption{Illustration of Offset Vectors.}
	\label{Illustration of Offset Vectors}
\end{figure}

This article 
designs a motion vector,
as illustrated in Fig. \ref{Illustration of Offset Vectors}. An offset vector $\boldsymbol{\delta}_t$ is introduced based on the fabric's edge endpoints to enable the robot to grasp points on the upper layer of the fabric. By adjusting 
$\boldsymbol{\delta}_t$, the robot can grasp any part of the fabric to manipulate it.

This article refines the operation process into three key steps: firstly, selecting one endpoint from the state variables as the base grasping point; secondly, generating an offset vector $\boldsymbol{\delta}_t$ to accurately adjust the position of the grasping point; thirdly, determining the coordinates of the placement point to guide the robot to complete the entire action from grasping to placing. The 
representation of the action space is described in (\ref{eq:action}):

\begin{equation}
	\boldsymbol{a}_t=(p_{\text{g}_t},\boldsymbol{\delta}_t,\boldsymbol{p}_{\text{p}_t}),
	\label{eq:action}
\end{equation}
where $p_{\text{g}_t}$ represents the index of the endpoint selected at time $t$ in the state variables, $\boldsymbol{\delta}_t$ represents the offset vector for time $t$, and $\boldsymbol{p}_{\text{p}_t}$ represents the coordinates of the placement point at time $t$.

\subsubsection{State Transition}
The state transition from time $t$ to time $t + 1$ is controlled by $\boldsymbol{a}_t$ as expressed in (\ref{eq:action}). Initially, the robot determines the coordinates of the corresponding endpoint $\boldsymbol{p}_{\text{g}_t}$ in $\boldsymbol{s}_t$ based on $p_{\text{g}_t}$. Subsequently, by combining $\boldsymbol{\delta}_t$, the actual grasping coordinates $\boldsymbol{g}_t=\boldsymbol{p}_{\text{g}_t}+\boldsymbol{\delta}_t$ are calculated, guiding the end effector to execute the grasping action at the $\boldsymbol{g}_t$ position. Finally, the robot moves the end effector to the $\boldsymbol{p}_{\text{p}_t}$ position, opens the gripper, and completes the placing operation.

\subsubsection{Reward Setting}
In the diagonal folding task, as shown in Fig. \ref{State Spaces for Different Tasks}, the target state of the fabric is when endpoint 1 and endpoint 3 coincide, and the distance between endpoint 2 and endpoint 4 equals the length of the diagonal, with the unfolded area of the fabric equal to half of its fully unfolded area. 
Assuming the side length of the square fabric is $l_s$, we define the difference $e_t$ between the fabric state at time $t$ and the target state in the diagonal folding task as follows:

\begin{equation}
	e_t=\big\|\boldsymbol{p}_{1_t}-\boldsymbol{p}_{3_t}\big\|_2+\Big|\sqrt{2}l_s-\big\|\boldsymbol{p}_{2_t}-\boldsymbol{p}_{4_t}\big\|_2\Big|+\big|f_t-0.5\big|.
	\label{eq:dia-fold-distance}
\end{equation}

The objective of folding along the central axis is to align endpoint 1 with endpoint 2, endpoint 3 with endpoint 4, ensure that the distance between endpoint 1 and endpoint 3 equals the distance between endpoint 2 and endpoint 4, both equal to ls, and the fabric's unfolded area equals half of its fully unfolded area. 
We define the gap $e_t$ between the fabric state at time $t$ and the target state in the folding along the central axis task as follows:

\begin{equation}\label{eq:middle-fold-distance}
	\begin{split}
		&e_t=\big\|\boldsymbol{p}_{1_t}-\boldsymbol{p}_{2_t}\big\|_2+\big\|\boldsymbol{p}_{3_t}-\boldsymbol{p}_{4_t}\big\|_2+\Big|l_s-\big\|\boldsymbol{p}_{1_t}-\boldsymbol{p}_{3_t}\big\|_2\Big|\\
		&+\Big|l_s-\big\|\boldsymbol{p}_{2_t}-\boldsymbol{p}_{4_t}\big\|_2\Big|+\big|f_t-0.5\big|.
	\end{split}
\end{equation}

For $e_t$ in different folding tasks, we define their reward functions as follows:

\begin{equation}
	r(\boldsymbol{s}_t,\boldsymbol{a}_t)=
	\begin{cases}
		-200e_t+100, & done \\
		3, & \text{not }done\text{ and } e_{t-1}-e_t>t_\text{z}\\
		-3, & \text{not }done\text{ and } e_t-e_{t-1}>t_\text{z}\\
		0, & \text{otherwise}
	\end{cases},
	\label{eq:fold-reward}
\end{equation}
where $done$ represents the completion status of the task, which becomes True when the maximum number of operations, $t_\text{m}$, is reached. $t_\text{z}$ is the threshold to measure whether the fabric state has significantly changed.
According to (\ref{eq:fold-reward}), the reward mechanism $r(\boldsymbol{s}_t,\boldsymbol{a}_t)$ assigns rewards or penalties to the agent based on its immediate actions and states, following the following guidelines:

\begin{itemize}
	\item{At the end of a round, a decisive reward of $-200e_t + 100$ is given based on the error $e_t$. The greater the error, the lower the decisive reward. The decisive reward is set to 100 when the $e_t$ is 0.}
	\item{If the round is not over and the error significantly decreases, i.e., $e_{t-1}-e_t>t_\text{z}$, indicating the agent is approaching the target, a positive reward of 3 is given to encourage similar behavior.}
	\item{Conversely, if the round is not over and the error significantly increases, i.e., $e_t-e_{t-1}>t_\text{z}$, indicating the agent deviates from the target, a negative penalty of -3 is applied to suppress this behavior.}
\end{itemize}

In the flattening task,  
we directly define the reward function for the flattening task based on the ratio $f_t$ of the fabric's unfolded area at time $t$ to its fully unfolded area:

\begin{equation}
	r(\boldsymbol{s}_t,\boldsymbol{a}_t)=
	\begin{cases}
		200f_t-100, & done \\
		3, & \text{not }done\text{ and } f_t-f_{t-1}>t_\text{z} \\
		-3, & \text{not }done\text{ and } f_{t-1}-f_t>t_\text{z}\\
		0, & \text{otherwise}
	\end{cases}.
	\label{eq:flatten-reward}
\end{equation}
The criteria followed here are similar to those described in (\ref{eq:fold-reward}), and will not be repeated here.

\subsection{Establishment and Analysis of the Spring-Mass Model}

We introduce a low-cost demonstration collection method based on NMPC. A NMPC problem is built based on the spring-mass particle model, and optimal control strategies are obtained by solving this problem to accomplish specific tasks. 
It achieves automated generation of demonstration data 
and significantly cuts data collection costs. 
However, in complex states of deformable objects (e.g., heavily wrinkled fabric), extracting the state of all particles 
in real environments poses significant challenges, limiting the feasibility of NMPC in real environments. Therefore, the purpose of the NMPC method is to automatically collect demonstration data in simulation to assist RL training.

\begin{figure}[tb]
	\centering
	\includegraphics[width=0.5\textwidth]{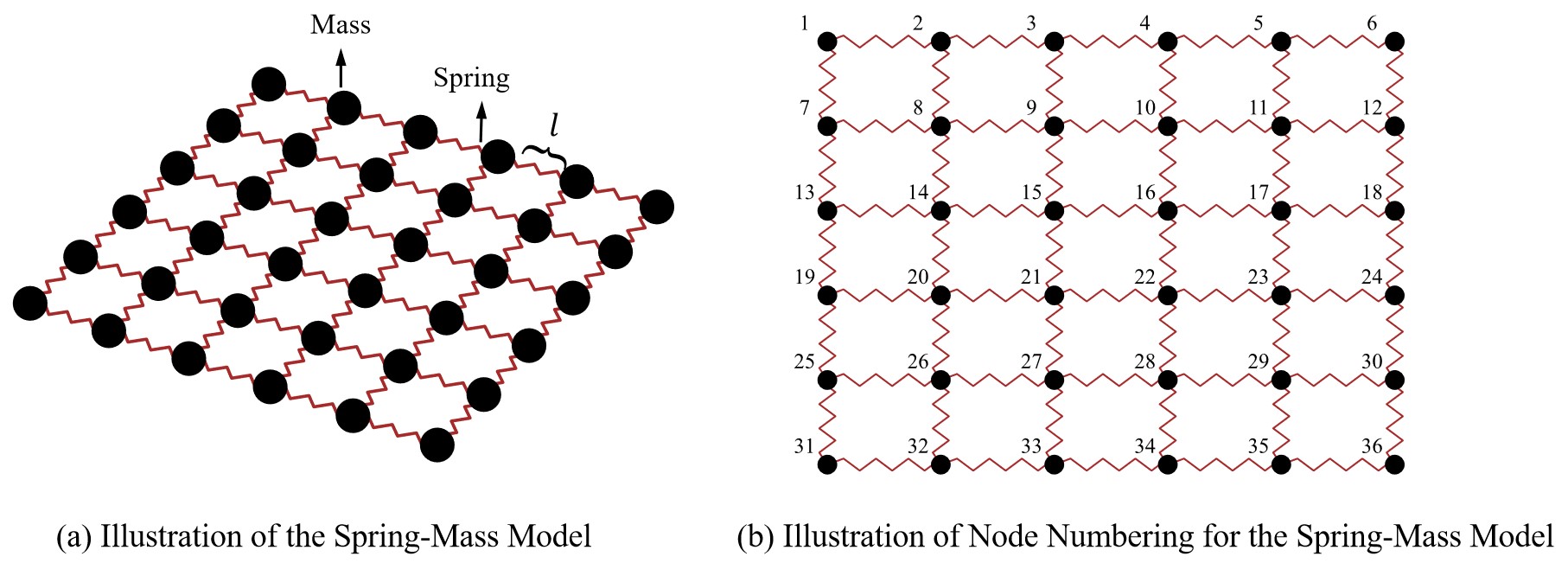}
	\caption{Illustration of the Spring-Mass Model and Node Numbering.}
	\label{Illustration of the Spring-Mass Model and Node Numbering}
\end{figure}

The spring-mass particle model adopted in this article is illustrated in Fig. \ref{Illustration of the Spring-Mass Model and Node Numbering}(a). This model can be viewed as a system of particles connected by multiple springs, with the mass of the cloth evenly distributed among the particles. The spring-mass particle model established in this article consists of a set of $N_{\text{sp}}$ particles denoted by $\mathcal{L}=(1,2,\cdots,N_{\text{sp}})$ and a set of $M$ springs denoted by $\mathcal{S}$. The set of neighbors of the $i$-th particle, i.e., the set of particles connected to particle $i$ by springs, is defined as $\mathcal{N}_i$, and it is assumed that the neighbor set of each particle is fixed.

We use a square cloth, so $N_{sp} = n^2$, where $n$ is the number of particles on one edge of the cloth. The mass of each particle $i$ is denoted by $m_i$, 
$m_i = m, \forall i$. The position of particle $i$ at time $t$ is represented by the three-dimensional coordinate vector $\boldsymbol{x}_t^i = (x_t^i, y_t^i, z_t^i)$. The stiffness coefficient of the springs usually depends on the physical properties of the material, while the natural length depends on the initial state of the cloth. We assume that all spring stiffness coefficients and natural lengths are equal, denoted as $k$ and $l$, respectively. The spring connecting particle $i$ and particle $j$ is denoted by $(i,j)$ (or $(j,i)$, which is equivalent). Thus, the neighbor set $\mathcal{N}_i$ can be expressed as:

\begin{equation}
	\mathcal{N}_i = \{j | (i,j)\in\mathcal{S}\}.\label{eq:neighbor_set}
\end{equation}

The particles are numbered from left to right and from top to bottom. Specifically, the particles in the first row are numbered from 1 to $n$, the particles in the second row are numbered from $n+1$ to $2n$, and so on. Fig. \ref{Illustration of the Spring-Mass Model and Node Numbering}(b) illustrates the case when $n=6$. 
Particles are initially subjected to the force exerted by the springs, which depends on the relative positions of the particles. For any two particles $i$ and $j$ connected by a spring, the spring force $\boldsymbol{s}_t^{i,j}$ at time $t$ can be expressed as:

\begin{equation}
	\boldsymbol{s}_t^{i,j} = k(l_t^{i,j} - l )\frac{\boldsymbol{x}_t^i - \boldsymbol{x}_t^j}{l_t^{i,j}},\label{eq:spring_force}
\end{equation}
where $l_t^{i,j}$ represents the actual distance between particles $i$ and $j$ at time $t$, which can be calculated using the following equation:

\begin{equation}
	l_t^{i,j}=\big\|\boldsymbol{x}_t^i - \boldsymbol{x}_t^j\big\|\label{eq:distance}
\end{equation}
The damping force 
can be expressed as follows:

\begin{equation}
	\boldsymbol{d}_t^{i,j} = -c(\boldsymbol{v}_t^i-\boldsymbol{v}_t^j),\label{eq:damping_force}
\end{equation}
where $c$ is the damping coefficient, $\boldsymbol{v}_t^i$ is the velocity vector of particle $i$ at time $t$, and $\boldsymbol{v}_t^i=(v_{t,x}^i, v_{t,y}^i, v_{t,z}^i)$, where $v_{t,x}^i, v_{t,y}^i, v_{t,z}^i$ are the velocity components of particle $i$ in the $x,y,z$ directions, respectively. Each particle is also influenced by gravity $\boldsymbol{G}=mg$, external force $\boldsymbol{u}_t^i$, and damping force $\boldsymbol{d}_t^{i,j}$. In summary, the total force $\boldsymbol{F}_t^i$ acting on particle $i$ at time $t$ can be expressed as:

\begin{equation}
	\boldsymbol{F}_t^i = \sum_{j\in\mathcal{N}_i} \boldsymbol{s}_t^{i,j} + \boldsymbol{G} + \boldsymbol{u}_t^i+\sum_{j\in\mathcal{N}_i}\boldsymbol{d}_t^{i,j}.\label{eq:total_force}
\end{equation}
We express the acceleration $\boldsymbol{a}_t^i$ of particle $i$ at time $t$ as:

\begin{equation}
	\boldsymbol{a}_t^i = \frac{\boldsymbol{F}_t^i}m.\label{eq:acceleration}
\end{equation}
Furthermore, the particle's velocity can be update as:

\begin{equation}
	\boldsymbol{v}_{t+\Delta t}^i = \boldsymbol{v}_t^i + \Delta t \cdot \boldsymbol{a}_t^i.\label{eq:velocity_update}
\end{equation}
Next, we treat $\boldsymbol{x}_t^i$ as a function of time $t$. Then, the position $\boldsymbol{x}_{t_0+\Delta t}^i$ of particle $i$ at time $t_0+\Delta t$ can be obtained by Second-order Taylor expanding at $t=t_0$:

\begin{equation}
	\boldsymbol{x}_{t_0+\Delta t}^i = \boldsymbol{x}_{t_0}^i + \Delta t \cdot \boldsymbol{v}_{t_0}^i + \frac{1}{2} \Delta t^2 \cdot \boldsymbol{a}_{t_0}^i.\label{eq:position_update}
\end{equation}
This article uses a small time step $\Delta t$ and sets the damping coefficient $c$ to a large positive value to ensure the stability of the model. In the following text, $t+n \cdot \Delta t$ is abbreviated as $t+n$ to simplify the subscript. The update formula for particle position is shown as (\ref{eq:position_update_final}):

\begin{equation}
	\boldsymbol{x}_{t+1}^i = \boldsymbol{x}_{t}^i + \Delta t \cdot \boldsymbol{v}_{t}^i + \frac{1}{2} \Delta t^2 \cdot \boldsymbol{a}_{t}^i.\label{eq:position_update_final}
\end{equation}

\section{Methodology}
\subsection{HGCR-DDPG Algorithm}
\subsubsection{Algorithm for Selecting Benchmark Grasping Points Based on the HTSK Fuzzy System}

In this article, as long as a suitable final $\boldsymbol{p}_{\text{p}_t}$ is chosen, selecting any reference $p_{\text{g}_t}$ can promote the task to some extent. Therefore, the selection of the fabric $p_{\text{g}_t}$ is more closely related to the application range of fuzzy sets.
We use the state-action pairs $(\boldsymbol{s}_t, \boldsymbol{a}_t)$, with the state $\boldsymbol{s}_t$ as input and $p_{\text{g}_t}$ from the action $\boldsymbol{a}_t$ as output, to construct an HTSK fuzzy system, denoted as $H(\boldsymbol{s}_t;\boldsymbol{\theta}^h)$, where $\boldsymbol{\theta}^h$ represents its parameters. The input-output relationship of $H(\boldsymbol{s}_t;\boldsymbol{\theta}^h)$ is represented as follows:

\begin{equation}
	p_{\text{g}_t} = H(\boldsymbol{s}_t;\boldsymbol{\theta}^h).
	\label{eq:HTSK}
\end{equation}

The training data for the selection strategy of the reference grasping point is sourced from the human demonstration dataset $\mathcal{D}_{\text{demo}}$. Additionally, data with significant contributions to task progress are continuously supplemented during the interaction between the robot and the environment, denoted as $\mathcal{D}_{\text{grasp}}=(\boldsymbol{s}_n,p_{\text{g}_n})^N_{n=1}$, where $N$ is the size of the training dataset, $\boldsymbol{s}_n=({s}_1^{(n)}, s_2^{(n)},\ldots,s_M^{(n)})$ represents the $M$-dimensional state variables of the $n$-th sample, $p_{\text{g}_n}\in (1,2,\cdots,k)$ is the index of the reference grasping point for the $n$-th sample, which is also the label of the training dataset, and $k$ is the number of candidate reference grasping points. The system learns the mapping relationship between the states $\boldsymbol{s}_n$ and the corresponding reference grasping points $p_{\text{g}_n}$ in the demonstration dataset, enabling it to predict appropriate reference grasping points under different states.
%


The primary improvement of HTSK over TSK is reflected in the saturation issues related to the dimensions of the data. In TSK fuzzy systems, the traditional softmax function is used to calculate the normalized firing levels $\overline{\omega}_r$ of each fuzzy rule, as follows:

\begin{equation}
	\overline{\omega}_r=\frac{\text{exp}(H_r)}{\sum_{r=1}^{R}\text{exp}(H_r)}.\label{eq:wr_tsk}
\end{equation}
Here, $H_r$ decreases as the dimension $M$ of the input data increases, leading to the saturation of the softmax function \cite{ca1}. In conventional TSK fuzzy systems, typically only the fuzzy rule with the highest $H_r$ receives a non-zero firing level $\overline{\omega}_r$. Consequently, as the data dimension $M$ increases, the distinctiveness of all $H_r$ values diminishes, and the classification performance of the TSK fuzzy system declines. To address this saturation issue, HTSK substitutes $H_r$ in the normalized firing levels $\overline{\omega}_r$ of each fuzzy rule within the TSK system with its mean value $H_r^*$, thereby allowing the normalization process to better accommodate high-dimensional data inputs. In the sixth and seventh layers of the HTSK net \cite{ca2}, based on the softmax function and the probability distribution, the final $p_{\text{g}_t}$ is selected as the output.

This article adopts the k-means clustering method to initialize $c_{r,m}$ which is the parameter of the Gaussian membership function, the cross-entropy loss function to measure the difference between the output of the fuzzy system and the true labels, and the Adam optimizer for gradient descent, with a learning rate of 0.04, a batch size of 64, and a weight decay of 1e-8.

\subsubsection{GABC-Improved Rainbow-DDPG Algorithm}

In Rainbow-DDPG, BC is typically implemented by adding $L_{\text{bc}}$ to the loss function of the Actor network, as shown in (\ref{eq:bcloss}):

\begin{equation}
	L_{\text{bc}}=\begin{cases}
		(\mu(\boldsymbol{s}_i;\boldsymbol{\theta})-\boldsymbol{a}_i)^2, & Q(\boldsymbol{s}_i,\boldsymbol{a}_i;\boldsymbol{\varphi})>Q(\boldsymbol{s}_i,\mu(\boldsymbol{s}_i;\boldsymbol{\theta});\boldsymbol{\varphi}) \\ & \text{ and }(\boldsymbol{s}_i,\boldsymbol{a}_i) \in \mathcal{D}_{\text{demo}}\\
		0, & \text{otherwise}
	\end{cases}.
	\label{eq:bcloss}
\end{equation}
The definition of $L_{\text{bc}}$ only takes effect when $Q(\boldsymbol{s}_i,\boldsymbol{a}_i;\boldsymbol{\varphi})>Q(\boldsymbol{s}_i,\mu(\boldsymbol{s}_i;\boldsymbol{\theta});\boldsymbol{\varphi})$. However, training the Critic network to output accurate Q-values is a time-consuming process, which results in the ineffectiveness of $L_{\text{bc}}$ in the early stages of training. Additionally, when the Critic network is fully trained, the replay buffer mainly contains real-time interaction data rather than demonstration data, reducing the probability of sampling demonstration data for training. Therefore, $L_{\text{bc}}$ may not have a significant impact, and RL still requires considerable training time to achieve good policies.

We propose GABC to improve Rainbow-DDPG for generating $\boldsymbol{\delta}_t$ and $\boldsymbol{p}_{\text{p}_t}$. We denote the current Actor network as $\mu(\boldsymbol{s}_t,p_{\text{g}_t};\boldsymbol{\theta}^{\mu})$, where $\boldsymbol{\theta}^{\mu}$ represents its parameters. In each state $\boldsymbol{s}_t$, this network combines with environmental noise $\mathcal{N}(0,\sigma^2)$ to output $\boldsymbol{\delta}_t$ and $\boldsymbol{p}_{\text{p}_t}$, guiding the robot to perform fabric manipulation tasks. The input-output relationship of $\mu(\boldsymbol{s}_t,p_{\text{g}_t};\boldsymbol{\theta}^{\mu})$ is represented as follows:

\begin{equation}
	(\boldsymbol{\delta}_t,\boldsymbol{p}_{\text{p}_t})=\mu(\boldsymbol{s}_t,p_{\text{g}_t};\boldsymbol{\theta}^{\mu}) + \mathcal{N}(0,\sigma^2).
\end{equation}
Its current Critic network is denoted as $Q(\boldsymbol{s}_t,\boldsymbol{a}_t;\boldsymbol{\theta}^q)$, where $\boldsymbol{\theta}^q$ represents its parameters. In each state $\boldsymbol{s}_t$, this network outputs a Q-value $Q(\boldsymbol{s}_t,\boldsymbol{a}_t;\boldsymbol{\theta}^q)$ through $\boldsymbol{\theta}^q$ to evaluate the quality of action $\boldsymbol{a}_t$. Since the quality of $\boldsymbol{\delta}_t$ and $\boldsymbol{p}_{\text{p}_t}$ is closely related to the selection of $p_{\text{g}_t}$, the input $\boldsymbol{a}_t$ of $Q(\boldsymbol{s}_t,\boldsymbol{a}_t;\boldsymbol{\theta}^q)$ not only includes $\boldsymbol{\delta}_t$ and $\boldsymbol{p}_{\text{p}_t}$ output by the Actor network but also includes $p_{\text{g}_t}$ output from demonstration data or the $H(\boldsymbol{s}_t;\boldsymbol{\theta}^h)$.

Assuming that during training, the sampled state-action data pairs are $(\boldsymbol{s}_i,\boldsymbol{a}_i)$, where $\boldsymbol{a}_i=(p_{\text{g}_i}, \boldsymbol{\delta}_i,\boldsymbol{p}_{\text{p}_i})$. This study denotes $(\boldsymbol{\delta}_i,\boldsymbol{p}_{\text{p}_i})$ as $\boldsymbol{b}_i$, and sets $\boldsymbol{w}_i=(p_{\text{g}_i},\mu(\boldsymbol{s}_i,p_{\text{g}_i};\boldsymbol{\theta}^{\mu}))$. Then, in the framework of this article, $L_{\text{bc}}$ can be redefined as follows:

\begin{equation}
	L_{\text{bc}}=\begin{cases}
		(\mu(\boldsymbol{s}_t,p_{\text{g}_t};\boldsymbol{\theta}^{\mu})-\boldsymbol{b}_i)^2, & Q(\boldsymbol{s}_i,\boldsymbol{a}_i;\boldsymbol{\theta}^q)>Q(\boldsymbol{s}_i,\boldsymbol{w}_i;\boldsymbol{\theta}^q) \\ & \text{ and }(\boldsymbol{s}_i,\boldsymbol{a}_i) \in \mathcal{D}_{\text{demo}}\\
		0, & \text{otherwise}
	\end{cases}.\label{eq:used_lbc}
\end{equation}

To expedite the training of the Critic, GABC introduces a loss term called $Q_{\text{diff}}$ into the current Critic network's loss function. Assuming $(\boldsymbol{s}_i,\boldsymbol{a}_i) \in \mathcal{D}_{\text{demo}}$, based on the fact that $\boldsymbol{\delta}_i$ and the placement point $\boldsymbol{p}_{\text{p}_i}$ in the human demonstration actions $\boldsymbol{a}_i$ are significantly superior to the offset vector and placement point output by the current Actor network $\mu(\boldsymbol{s}_i,p_{\text{g}_i};\boldsymbol{\theta}^{\mu})$, $Q_{\text{diff}}$ guides the training of the Critic network by measuring the difference between the $Q$-values output by the current Critic network for the actions $\boldsymbol{a}_i$ in the human demonstration data and the $Q$-values output for the actions $\boldsymbol{w}_i=(p_{\text{g}_i},\mu(\boldsymbol{s}_i,p_{\text{g}_i};\boldsymbol{\theta}^{\mu}))$ by the current Actor network, under the same state $\boldsymbol{s}_i$. Specifically, this article sets a pre-training stage, during which, in the pre-training phase, when $Q(\boldsymbol{s}_i,\boldsymbol{a}_i;\boldsymbol{\theta}^q)-Q(\boldsymbol{s}_i,\boldsymbol{w}_i;\boldsymbol{\theta}^q)\geq100$, $Q_{\text{diff}}$ is set to 0; otherwise, $Q_{\text{diff}}=100-(Q(\boldsymbol{s}_i,\boldsymbol{a}_i;\boldsymbol{\theta}^q)-Q(\boldsymbol{s}_i,\boldsymbol{w}_i;\boldsymbol{\theta}^q))$, as shown in (\ref{eq:qdiff}):

\begin{equation}
	Q_{\text{diff}}=\max(0,100-(Q(\boldsymbol{s}_i,\boldsymbol{a}_i;\boldsymbol{\theta}^q)-Q(\boldsymbol{s}_i,\boldsymbol{w}_i;\boldsymbol{\theta}^q))).
	\label{eq:qdiff}
\end{equation}

After pre-training is completed, 
the Actor network has already acquired a certain policy, and its output actions $\mu(\boldsymbol{s}_i,p_{\text{g}_i};\boldsymbol{\theta}^{\mu})$ may not significantly inferior to the actions $\boldsymbol{a}_i$ in the human demonstration data. The introduction of $Q_{\text{diff}}$ may lead to the training of the Actor network getting stuck in local optima, 
so that the $Q_{\text{diff}}$ is removed.

The loss function $L_{\text{Critic}}$ for the improved Critic network $Q(\boldsymbol{s}_t,\boldsymbol{a}_t;\boldsymbol{\theta}^q)$ is defined as follows: 

\begin{equation}
	L_{\text{Critic}}=\lambda_{\text{1step}}L_{\text{1s}}+\lambda_{n\text{step}}L_{n\text{s}}+\lambda_{\text{diff}}Q_{\text{diff}},
	\label{eq:Criticloss}
\end{equation}
where $\lambda_{\text{1step}}$ and $\lambda_{n\text{step}}$ are the weights of the 1-step and n-step TD loss functions, respectively. $\lambda_{\text{diff}}$ is the weight of $Q_{\text{diff}}$, set to 1 during the pre-training phase and 0 in subsequent phases. $L_{\text{1s}}$ and $L_{n\text{s}}$ are similar to the 1-step and n-step TD loss functions \cite{c18}, and the target functions $y_{\text{1s}}$ and $y_{n\text{s}}$ can be referenced from TD3 model \cite{c28}. Here, we define $Q^\prime_1$, $Q^\prime_2$, and $\mu^\prime$ as two target Critic networks and one target Actor network, with $\boldsymbol{w}^\prime_{i+k}=(H(\boldsymbol{s}_{i+k};\boldsymbol{\theta}^h),\mu^\prime(\boldsymbol{s}_{i+k},H(\boldsymbol{s}_{i+k};\boldsymbol{\theta}^h);\boldsymbol{\theta}^{{\mu}^\prime})),k=1,2,\cdots,n,$ where $n$ is the step length of the n-step TD loss function, $N$ is the batch size, $r_i$ is the reward, and $\gamma$ is the discount factor. Consequently, $L_{\text{1s}}$ and $L_{\text{ns}}$ are defined as follows:

\begin{equation}
	L_{\text{1s}}=\frac{1}{N}\sum_{i=1}^{N}(Q(\boldsymbol{s}_i,\boldsymbol{a}_i;\boldsymbol{\theta}^q)-y_{\text{1s}})^2,
\end{equation}
\begin{equation}
	y_{\text{1s}}=
	r_i+\gamma\min_{j=1,2} Q^\prime_j(\boldsymbol{s}_{i+1},\boldsymbol{w}^\prime_{i+1};\boldsymbol{\theta}^{q^\prime}),
\end{equation}
\begin{equation}
	L_{n\text{s}}=\frac{1}{N}\sum_{i=1}^{N}(Q(\boldsymbol{s}_i,\boldsymbol{a}_i;\boldsymbol{\theta}^q)-y_{n\text{s}})^2,
\end{equation}
\begin{equation}
	y_{n\text{s}}=\sum_{t=0}^{n-1}{\gamma^tr_{i+t+1}+\gamma^n\min_{j=1,2}Q^\prime_j(\boldsymbol{s}_{i+n},\boldsymbol{w}^\prime_{i+n};\boldsymbol{\theta}^{{q}^\prime})}.
\end{equation}
During training, the introduction of $Q_{\text{diff}}$ causes the current Critic network to initially tend towards generating larger $Q$-values for actions $\boldsymbol{a}_i$ from human demonstrations,
the effect of $L_{\text{bc}}$ in (\ref{eq:used_lbc}) becomes more pronounced, leading to a more thorough utilization of human demonstrations and thus accelerating the training of the Actor network. Additionally, the training of the current Actor network $\mu(\boldsymbol{s}_i,p_{\text{g}_i};\boldsymbol{\theta}^{\mu})$ also depends on the $Q$-values output by the current Critic network $\mu(\boldsymbol{s}_t,p_{\text{g}_t};\boldsymbol{\theta}^{\mu})$. Consequently, a current Critic network capable of providing more precise $Q$-values will further accelerate the training of the current Actor network.

\subsubsection{CPL Learning Method}

\begin{figure}[tb]
	\centering
	\includegraphics[width=0.35\textwidth]{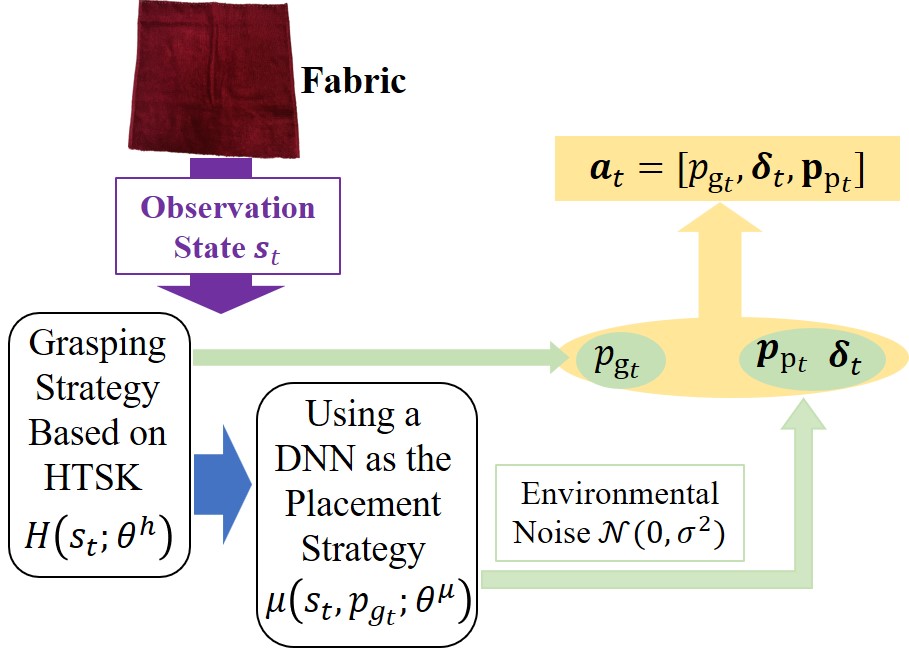}
	\caption{Illustration of Conditional Policy Learning.}
	\label{Illustration of Conditional Policy Learning}
\end{figure}

This article adopts the CPL training method illustrated in Fig. \ref{Illustration of Conditional Policy Learning}. At each state $\boldsymbol{s}_t$, CPL first utilizes the $H(\boldsymbol{s}_t;\boldsymbol{\theta}^h)$ to select $p_{\text{g}_t}$, and then, based on $p_{\text{g}_t}$, selects $\boldsymbol{\delta}_t$ and $\boldsymbol{p}_{\text{p}_t}$ through $\mu(\boldsymbol{s}_t,p_{\text{g}_t};\boldsymbol{\theta}^{\mu})$.

CPL often faces the challenge of loss allocation \cite{c27}. It's difficult to determine whether high rewards obtained for an action are attributed to the grasping policy $H(\boldsymbol{s}_t;\boldsymbol{\theta}^h)$ or the offset vector and placement point selection policy $\mu(\boldsymbol{s}_t,p_{\text{g}_t};\boldsymbol{\theta}^{\mu})$. 
To address this, the following training approach is adopted. Firstly, each state-action pair $(\boldsymbol{s}_t,\boldsymbol{a}_t)$ is extracted from $\mathcal{D}_{\text{demo}}$ to form an initial training dataset $\mathcal{D}_{\text{grasp}}$ tailored for the HTSK fuzzy system. Then, $\mathcal{D}_{\text{grasp}}$ is used to train $H(\boldsymbol{s}_t;\boldsymbol{\theta}^h)$ to get $\boldsymbol{\theta}^h$. Subsequently, with $\boldsymbol{\theta}^h$ fixed, based on the $H(\boldsymbol{s}_t;\boldsymbol{\theta}^h)$ to select $p_{\text{g}_t}$, the improved Rainbow-DDPG algorithm with GABC enhancement is employed to train the offset vector and placement point selection policy $\mu(\boldsymbol{s}_t,p_{\text{g}_t};\boldsymbol{\theta}^{\mu})$, obtaining $\boldsymbol{\theta}^{\mu}$. 

During the process, data is also collected to supplement the training dataset $\mathcal{D}_{\text{grasp}}$ and continuously train the improved grasping policy $H(\boldsymbol{s}_t;\boldsymbol{\theta}^h)$ parameters $\boldsymbol{\theta}^h$. 
Specifically, when the action $\boldsymbol{a}_t$ executed by the robot at time $t$ significantly advances the progress of the task (in folding tasks, $e_{t-1}-e_{t}>t_\text{z}$, in flattening tasks, $f_{t}-f_{t-1}>t_\text{z}$), $(\boldsymbol{s}_t,\boldsymbol{a}_t)$ is added to $\mathcal{D}_{\text{grasp}}$. After getting a certain amount of new data, $\mathcal{D}_{\text{grasp}}$ is used to retrain the grasping policy $H(\boldsymbol{s}_t;\boldsymbol{\theta}^h)$ to get new parameters $\boldsymbol{\theta}^h$. 

Combining all the improvements introduced above, we proposed the HGCR-DDPG algorithm. The relevant pseudocode is detailed in Algorithm \ref{alg:condition-policy-learning}.

\begin{algorithm}[tb]
	\caption{HGCR-DDPG} 
	\label{alg:condition-policy-learning} 
	\begin{algorithmic}[1] 
		\REQUIRE Demonstration Dataset $\mathcal{D}_{\text{demo}}$, Total Number of Rounds$M$, Number of Pre-Training Rounds$M_p$, Maximum Number of Operations per Round$t_{\text{m}}$, Number of Strategy Updates per Interaction$t_\text{n}$, Batch Size $N$, Random Environmental Noise$\mathcal{N}$.
		\ENSURE Trained HTSK Fuzzy System $H(\boldsymbol{s}_t;\boldsymbol{\theta}^h)$, Critic Network
		$Q(\boldsymbol{s}_t,\boldsymbol{a}_t;\boldsymbol{\theta}^q)$ and Offset Vector \& Placement Point Selection Strategy Network $\mu(\boldsymbol{s}_t,p_{\text{g}_t};\boldsymbol{\theta}^{\mu})$.
		\STATE Extract$(\boldsymbol{s}_i,\boldsymbol{a}_i)$ from $\mathcal{D}_{\text{demo}}$to form $\mathcal{D}_{\text{grasp}}$. Add $\mathcal{D}_{\text{demo}}$ to the replay buffer R. Initialize $H(\boldsymbol{s}_t;\boldsymbol{\theta}^h)$, $Q(\boldsymbol{s}_t,\boldsymbol{a}_t;\boldsymbol{\theta}^q)$, $\mu(\boldsymbol{s}_t,p_{\text{g}_t};\boldsymbol{\theta}^{\mu})$, $Q'_1(\boldsymbol{s}_t,\boldsymbol{a}_t;\boldsymbol{\theta}^{q'})$, $Q'_2(\boldsymbol{s}_t,\boldsymbol{a}_t;\boldsymbol{\theta}^{q'})$, $\mu'(\boldsymbol{s}_t,p_{\text{g}_t};\boldsymbol{\theta}^{\mu'})$.             
		\WHILE{$H(\boldsymbol{s}_t;\boldsymbol{\theta}^h)$has not converged}
		\STATE Use $\mathcal{D}_{\text{grasp}}$ as the training dataset, and update the parameters $\boldsymbol{\theta}^h$ of  $H(\boldsymbol{s}_t;\boldsymbol{\theta}^h)$ according to the method described in Section II.B.1).
		\ENDWHILE
		\FOR{$e=1,M$}
		\STATE Initialize the environment and receive the initial observation state $\boldsymbol{s}_1$.
		\FOR{$t$ = 1,$t_{\text{m}}$}
		\STATE Referring to Fig. \ref{Illustration of Conditional Policy Learning}, generate $\boldsymbol{a}_t$ by combining $H(\boldsymbol{s}_t;\boldsymbol{\theta}^h)$, $\mu(\boldsymbol{s}_t,p_{\text{g}_t};\boldsymbol{\theta}^{\mu})$, and noise $\mathcal{N}$.
		\STATE Execute action $\boldsymbol{a}_t$to interact with the environment, and add $(\boldsymbol{s}_t,\boldsymbol{a}_t,r_t,\boldsymbol{s}_{t+1})$ to the replay buffer R.
		\FOR{$b=1,t_\text{n}$}
		\IF{$e \leq M_p$}
		\STATE Set $\lambda_{\text{diff}}$ to 1, and sample $N$ data points $(\boldsymbol{s}_i,\boldsymbol{a}_i,r_i,\boldsymbol{s}_{i+1})$ from $\mathcal{D}_{\text{demo}}$.
		\ELSE
		\STATE Set $\lambda_{\text{diff}}$ to 0, and sample $N$ data points $(\boldsymbol{s}_i,\boldsymbol{a}_i,r_i,\boldsymbol{s}_{i+1})$ from R.
		\ENDIF
		\STATE Update $Q(\boldsymbol{s}_t,\boldsymbol{a}_t;\boldsymbol{\theta}^q)$ by minimizing the loss $L_{\text{Critic}}$ defined in (\ref{eq:Criticloss}).
		\STATE Update $\mu(\boldsymbol{s}_t,p_{\text{g}_t};\boldsymbol{\theta}^{\mu})$ using the method in Rainbow-DDPG.
		\STATE If $\boldsymbol{a}_i$significantly advances the task, add $(\boldsymbol{s}_i,\boldsymbol{a}_i)$ to $\mathcal{D}_{\text{grasp}}$. If there are more than 50 new data points added to $\mathcal{D}_{\text{grasp}}$, retrain $H(\boldsymbol{s}_t;\boldsymbol{\theta}^h)$.
		\ENDFOR
		\STATE Soft-update the parameters of the target networks $Q'(\boldsymbol{s}_t,\boldsymbol{a}_t;\boldsymbol{\theta}^{q'})$ and $\mu'(\boldsymbol{s}_t,p_{\text{g}_t};\boldsymbol{\theta}^{\mu'})$.
		\ENDFOR
		\ENDFOR
	\end{algorithmic}
\end{algorithm}

\subsection{Low-Cost Demonstration Collection Based on NMPC}

In this article, the objective of NMPC control is to find an optimal control sequence $\boldsymbol{U}_{list}^* = (\boldsymbol{U}_t^*, \boldsymbol{U}_{t+1}^*, \cdots, \boldsymbol{U}_{t+H_p-1}^*)$ within the prediction horizon $H_p$ to minimize the objective function $J$. The system's state consists of the coordinates of each particle $\boldsymbol{X}_t=(\boldsymbol{x}_t^1,\boldsymbol{x}_t^2,\cdots,\boldsymbol{x}_t^{N_{sp}})$, and the control inputs are the external forces applied to each particle $\boldsymbol{U}_t=(\boldsymbol{u}_t^1,\boldsymbol{u}_t^2,\cdots,\boldsymbol{u}_t^{N_{sp}})$, where $\boldsymbol{u}_t^i=(u_{t,x}^i, u_{t,y}^i, u_{t,z}^i)$, $u_{t,x}^i, u_{t,y}^i, u_{t,z}^i$ are the components of the external force in the $x,y,z$ directions, respectively. Consequently, the state transition equation for the spring-mass model can be expressed as:

\begin{equation}
	\begin{aligned}
		\boldsymbol{X}_{t+1}&=\boldsymbol{f}(\boldsymbol{X}_{t}, \boldsymbol{U}_t)\\
		&=\boldsymbol{X}_{t} + \Delta t \cdot \boldsymbol{V}_{t} + \frac{1}{2 \cdot m} \Delta t^2 \cdot (\boldsymbol{S}_{t} + \boldsymbol{G} + \boldsymbol{U}_t + \boldsymbol{D}_t)
	\end{aligned},
	\label{eq:state_transition}
\end{equation}
where $\boldsymbol{V}_{t}=(\boldsymbol{v}_t^1,\boldsymbol{v}_t^2,\cdots,\boldsymbol{v}_t^{N_{sp}})$ is the velocity of each particle, $\boldsymbol{S}_{t}=(\boldsymbol{s}_t^1,\boldsymbol{s}_t^2,\cdots,\boldsymbol{s}_t^{N_{sp}})$ is the spring force applied to each particle, $\boldsymbol{G}$ is the gravity acting on each particle, $\boldsymbol{D}_{t}=(\boldsymbol{d}_t^1,\boldsymbol{d}_t^2,\cdots,\boldsymbol{d}_t^{N_{sp}})$ is the damping force applied to each particle, and $m$ is the mass of each particle.

The design of the loss function is based on the distances between particles. 
Specifically, for the three task objectives, each of which is specified by the distances $l^{i,j}_t$ between particles at time $t$. Taking the particle ordering in Fig. \ref{Illustration of the Spring-Mass Model and Node Numbering} as an example, the target state $\boldsymbol{X}_{\text{ref}}$ is redefined as follows:

1. Folding along the diagonal: 
For any pair of particles $i$ and $j$ symmetrically positioned about the specified diagonal, in the target state, the distance $l^{i,j}_t$ between them should satisfy:

\begin{equation}
	l^{i,j}_t = 0,\text{ such as }l^{1,36}_t = l^{8,29}_t = \ldots = 0.
	\label{eq:l_folding_diagonal}
\end{equation}

2. Folding along the central axis: 
For any pair of particles $i$ and $j$ symmetrically positioned about the specified central axis, in the target state, the distance $l^{i,j}_t$ between them should satisfy:

\begin{equation}
	l^{i,j}_t = 0,\text{ such as }l^{1,6}_t = l^{8,11}_t = \ldots = 0.
	\label{eq:l_folding_central_axis}
\end{equation}

3. Flattening: 
For the particles $i$ and $j$ at the ends of the two diagonals of the cloth, and for the particles $a$ and $b$, in the target state, the distances between them should satisfy:

\begin{equation}
	l^{i,j}_t = l^{a,b}_t = \sqrt{2}l_s,\text{ such as }l^{1,36}_t = l^{6,31}_t = \sqrt{2}l_s,
	\label{eq:l_flattening}
\end{equation}
where $l_s$ is the side length of the cloth in the fully flattened state.

For a given task objective, the loss function $L(\boldsymbol{X}_t)$ can be defined as:

\begin{equation}
	L(\boldsymbol{X}_t) = \sum_{i,j \in \mathcal{P}} w_{i,j} ( l^{i,j}_t - l^{i,j}_{\text{ref}} )^2,
\end{equation}
where $\mathcal{P}$ is the set of all pairs of particles to be considered, $l^{i,j}_{\text{ref}}$ is the desired distance between particles $i$ and $j$ in the target state $\boldsymbol{X}_{\text{ref}}$, \( w_{i,j} \) is the weight factor used to adjust the relative importance of the differences in distances between different pairs of particles.
For a specific task objective, the value of $l^{i,j}_{\text{ref}}$ can be chosen based on (\ref{eq:l_folding_diagonal}) to (\ref{eq:l_flattening}).
%
%

The objective of NMPC is to minimize the cumulative loss within the prediction horizon. (\ref{eq:state_transition}) is used to predict future states based on the initial state and control inputs:

\begin{equation}
	\boldsymbol{X}_{k+1|t} = \boldsymbol{f}(\boldsymbol{X}_{k|t}, \boldsymbol{U}_k), \quad  k=t,\ldots,t+H_p-1,
\end{equation}
where $\boldsymbol{X}_{k|t}$ and $\boldsymbol{X}_{k+1|t}$ represent the states at time steps $k$ and $k + 1$ predicted based on the current state $\boldsymbol{X}_t$ and a series of control inputs $\boldsymbol{U}_{list}$. Specifically, when $k=t$, $\boldsymbol{X}_{k|t}=\boldsymbol{X}_{t|t}=\boldsymbol{X}_{t}$.

In summary, the NMPC in this article can be implemented by solving the following optimization problem in (\ref{eq:nmpc_optimization_problem}):
\begin{equation}
	\begin{aligned}
		& \underset{\boldsymbol{U}_{list}}{\text{minimize}}
		& & J(\boldsymbol{X}_t, \boldsymbol{U}_{list}) = \sum_{k=t+1}^{t+H_p}L(\boldsymbol{X}_{k|t})\\& & &=\sum_{k=t+1}^{t+H_p} \sum_{i,j \in \mathcal{P}} w_{i,j} ( l^{i,j}_{k|t} - l^{i,j}_{\text{ref}} )^2 \\
		& \text{subject to}
		& & \boldsymbol{X}_{k+1|t} = \boldsymbol{f}(\boldsymbol{X}_{k|t}, \boldsymbol{U}_k) \\
		&&& \quad  k=t,\ldots,t+H_p-1\\
		&&& \boldsymbol{X}_{t|t}=\boldsymbol{X}_{t} \\
		&&& -10N \leq u_{k,x}^i, u_{k,y}^i, u_{k,z}^i \leq 10N, \\
		&&&\text{      }\quad (i=1,\ldots,N, \; k=t,\ldots,t+H_p-1), \\
		&&& \boldsymbol{x}_{k|t}^i \in \mathcal{X}_{\text{workspace}}, \\
		&&&\text{      }\quad (i=1,\ldots,N, \; k=t+1,\ldots,t+H_p),
	\end{aligned}
	\label{eq:nmpc_optimization_problem}
\end{equation}
where $J(\boldsymbol{X}_t, \boldsymbol{U}_{list})$ represents the cumulative loss function over the entire prediction horizon $H_p$. The term $l^{i,j}_{k|t}$ denotes the distance between the $i$-th and $j$-th particles calculated based on $\boldsymbol{X}_{k|t}$ at the $k$-th time step. $\mathcal{X}_{\text{workspace}}$ denotes the set of state constraints in the robot's workspace. We utilized the Interior Point OPTimizer (Ipopt), which is based on interior-point methods for nonlinear programming \cite{c29}. By solving the optimization problem described above, we can obtain the optimal control sequence $\boldsymbol{U}_{list}^*$ that minimizes the loss function within the prediction horizon.

\begin{figure}[tb]
	\centering
	\includegraphics[width=0.4\textwidth]{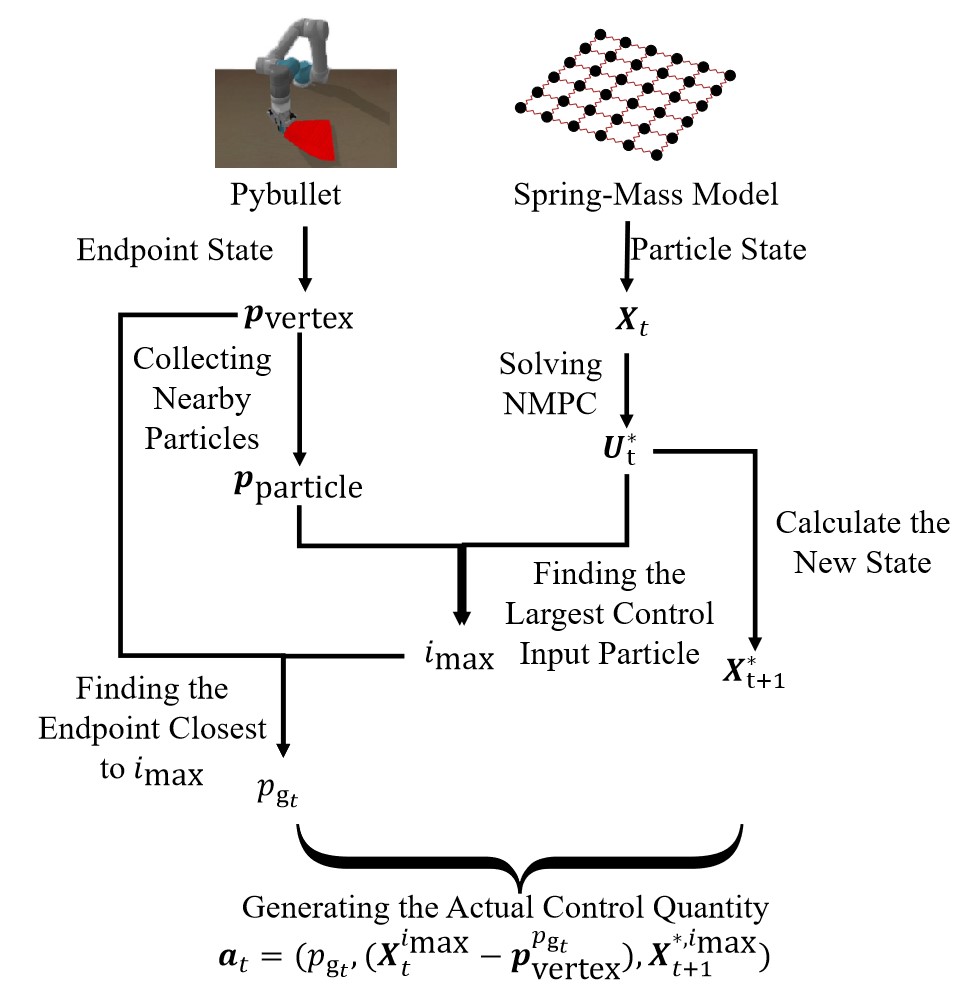}
	\caption{Process for Generating the Robot's Motion Space Based on NMPC.}
	\label{Process for Generating the Robot's Motion Space Based on NMPC}
\end{figure}

The optimal control sequence $\boldsymbol{U}_{t}^*$ obtained from solving the NMPC problem defines the ideal external forces applied to each particle of the system at time $t$. However, the robot can only apply force to a single particle at any given time. Therefore, to translate NMPC into a practically executable robot control strategy, $\boldsymbol{U}_{t}^*$ must be mapped to the robot's action space. The specific process is illustrated in Fig. \ref{Process for Generating the Robot's Motion Space Based on NMPC}, and the detailed explanation follows.

Firstly, the optimal control sequence $\boldsymbol{U}_{t}^*$ is applied to the dynamic equations of the spring-mass particle model to predict the system's state $\boldsymbol{X}_{t+1}^*$ at the next time step:

\begin{equation}
	\boldsymbol{X}_{t+1}^* = \boldsymbol{f}(\boldsymbol{X}_{t}, \boldsymbol{U}_t^*).
\end{equation}

Next, utilizing the PyBullet simulation environment, we obtain the current state vector of the system, from which we extract the coordinates of the model's endpoints, denoted as $\boldsymbol{P}_{\text{vertex}}=(\boldsymbol{p}_{\text{vertex}}^1,\boldsymbol{p}_{\text{vertex}}^2, \cdots,\boldsymbol{p}_{\text{vertex}}^k)$, where $k$ is the number of endpoints in the state vector.

Subsequently, we analyze each endpoint $\boldsymbol{p}_{\text{vertex}}^i$, identify the nearest 10 particles and the neighboring particles of these identified particles to form the set $\boldsymbol{p}_{\text{particle}}^i$. 
Therefore, for all endpoints, we construct a broader set $\boldsymbol{P}_{\text{particle}}=(\boldsymbol{p}_{\text{particle}}^1, \cdots, \boldsymbol{p}_{\text{particle}}^k)$.

By analyzing $\boldsymbol{U}_t^*$, we identify the maximum external force $\boldsymbol{u}_{t}^{*,i_{\text{max}}}$ acting on all particles in $\boldsymbol{P}_{\text{particle}}$, and determine the index $p_{\text{g}_t}$ of the endpoint nearest to particle $i_{\text{max}}$in the state vector $\boldsymbol{s}_t$ as the reference grasping point for the robot. We calculate the displacement vector between particle $i_{\text{max}}$ and endpoint $p_{\text{g}_t}$ as the grasping offset vector $\boldsymbol{\delta}_t$.

Finally, based on the predicted position information $\boldsymbol{X}_{t+1}^*$, we determine the ideal position $\boldsymbol{x}_{t+1}^{*,i_{\text{max}}}$ of particle $i_{\text{max}}$ at the next time step and designate it as the placement point coordinate $\boldsymbol{p}_{\text{p}_t}$, completing the conversion from theoretical control quantities to actual robot motion commands.

In summary, for the optimal control quantity $\boldsymbol{U}_{t}^*$ computed by NMPC, we transform it to the action space:

\begin{equation}
	\boldsymbol{a}_t=(p_{\text{g}_t},(\boldsymbol{x}_{t}^{i_{\text{max}}}-\boldsymbol{p}_{\text{vertex}}^{p_{\text{g}_t}}),\boldsymbol{x}_{t+1}^{*,i_{\text{max}}}).\label{eq:action_nmpc}
\end{equation}

Additionally, due to potential simulation errors in the spring-mass particle model, 
we utilize the PyBullet simulation environment to correct the errors in the spring-mass particle model at each time step, as illustrated in Fig. S3. 
After obtaining $\boldsymbol{a}_t$ from NMPC as in (\ref{eq:action_nmpc}), it is executed in PyBullet to obtain the new state of the cloth, which is then used as the initial state $\boldsymbol{X}_{t+1}$ for the next control cycle, thereby achieving precise updating of the cloth state. 

We use $dn$ to denote whether a demonstration episode has ended. The termination condition for an episode is $|e_t-e_{t-1}| \leq 0.01$ for folding tasks or $|f_t-f_{t-1}| \leq 0.01$ for flattening tasks. 
We collect episodes with relatively high rewards as demonstration data, resulting in the NMPC demonstration dataset $\mathcal{D}_{\text{NMPC}}$, as shown in Algorithm \ref{alg:nmpc}.

\begin{algorithm}[tb]
	\caption{NMPC Demonstration Data Collection} 
	\label{alg:nmpc} 
	\begin{algorithmic}[1] 
		\REQUIRE System model $\boldsymbol{f}(\boldsymbol{X}_t,\boldsymbol{U}_t)$, initial state $\boldsymbol{X}_0$, prediction horizon $H_p$, cost function $J(\boldsymbol{X}_t, \boldsymbol{U}_{list})$, reward threshold $r_{\text{ts}}$ for each task.
		\ENSURE NMPC demonstration dataset $\mathcal{D}_{\text{NMPC}}$.
		\STATE Initialize temporary dataset $\mathcal{D}_{\text{temp}}$, set $dn$ to False.
		\WHILE {sufficient demonstration data has not been collected}
		\STATE Formulate the NMPC optimization problem according to (\ref{eq:nmpc_optimization_problem}).
		\WHILE {$\text{not } dn$}
		\STATE Solve the optimization problem using Ipopt to obtain the control sequence $\boldsymbol{U}_{list}^*=(\boldsymbol{U}_t^*, \boldsymbol{U}_{t+1}^*, \cdots, \boldsymbol{U}_{t+H_p-1}^*)$.
		\STATE Extract $\boldsymbol{U}_t^*$ and generate the action vector $\boldsymbol{a}_t$ according to (\ref{eq:action_nmpc}).
		\STATE Apply $\boldsymbol{a}_t$ to the PyBullet simulation environment to control the robot and update the fabric state.
		\STATE Retrieve the new fabric state $\boldsymbol{X}_{t+1}$, as well as the state vector $\boldsymbol{s}_{t+1}$, reward $r_{t}$, and done flag $dn$ from the PyBullet simulation environment.
		\STATE Store $(\boldsymbol{s}_{t},\boldsymbol{a}_t,r_t,\boldsymbol{s}_{t+1})$ in $\mathcal{D}_{\text{temp}}$.
		\STATE Set $\boldsymbol{X}_{t+1}$ as the initial state for the next control cycle.
		\ENDWHILE
		\IF {$r_t \geq r_{\text{ts}}$}
		\STATE Add $\mathcal{D}_{\text{temp}}$ to $\mathcal{D}_{\text{NMPC}}$.
		\ENDIF
		\STATE Clear $\mathcal{D}_{\text{temp}}$ and reset the simulation environment for the next control task.
		\ENDWHILE
	\end{algorithmic}
\end{algorithm}

%

\section{Experimental Setup}
\subsection{Simulation Experiment Settings of the Improved HGCR-DDPG Algorithm with Human Demonstrations}

As shown in Fig. S4, in simulation, humans guide the robot to perform grasping and placing actions by clicking on the grasp point $\boldsymbol{p}_{\text{gr}_{t}}$ and the placement point $\boldsymbol{p}_{\text{p}_t}$ with a mouse, respectively. During this process, we identify the endpoint closest to $\boldsymbol{p}_{\text{gr}_{t}}$ in state $\boldsymbol{s}_t$, denote its coordinates as $\boldsymbol{p}_{\text{g}_{t}}$, and its index as $p_{\text{g}_t}$, and calculate the offset vector as $\boldsymbol{\delta}_t=\boldsymbol{p}_{\text{gr}_{t}}-\boldsymbol{p}_{\text{g}_{t}}$. Subsequently, we obtain the action $\boldsymbol{a}_t=(p_{\text{g}_t},\boldsymbol{\delta}_t,\boldsymbol{p}_{\text{p}_t})$, and organize it along with state, reward, and other information into a tuple $(\boldsymbol{s}_t,\boldsymbol{a}_t,r_t,\boldsymbol{s}_{t+1})$. At the end of each demonstration, all tuples of the round are stored in a dedicated dataset $\mathcal{D}_{\text{demo}}$ to assist in training the DRL. During data collection, the end condition for rounds of both folding tasks is $e_t \leq 0.1$, and for flattening tasks, the end condition for rounds is $f_t \geq 0.9$. 

In Table. \ref{tab:Human Demonstration Dataset}, we present the analysis results of human demonstration datasets for three different tasks (diagonal folding, axial folding, and flattening). 
For the diagonal folding task, the demonstrators achieve the highest average reward (93.662),  the most stable performance (with a reward standard deviation of only 3.105), and the fewest rounds (1.000) to successfully complete the task. While for the axial folding task, the average reward is 87.741, the reward standard deviation is 3.624, and the number of rounds needed to complete the task is 2.868. These metrics indicate that although the quality of task execution remains relatively high, consistency and efficiency have slightly decreased compared to diagonal folding. In the flattening task, although the average reward (87.688) is comparable to that of axial folding, the standard deviation of the reward significantly increased to 5.572, and the number of steps required to complete the task surged to 8.291, indicating the high complexity of the flattening task. 

\begin{table}[tb]
	\centering
	\caption{Human Demonstration Dataset}
	\begin{tabular}{c|ccc}
		\toprule
		\diagbox [width=6em,trim=l] {Task}{Metric} & \begin{tabular}[c]{@{}c@{}}
			Average\\Reward
		\end{tabular} & \begin{tabular}[c]{@{}c@{}}
		Reward\\Standard Deviation
	\end{tabular} & \begin{tabular}[c]{@{}c@{}}
	Average\\Steps
\end{tabular} \\
		\hline
		\begin{tabular}[c]{@{}c@{}}
			Folding Along\\the Diagonal
		\end{tabular} & 93.662 & 3.105 & 1.000 \\
			\begin{tabular}[c]{@{}c@{}}
			Folding Along\\the Central Axis
		\end{tabular}& 87.741 & 3.624 & 2.868 \\
		Flattening& 87.688 & 5.572 & 8.291 \\
		\bottomrule
		\hline
	\end{tabular}%
	\label{tab:Human Demonstration Dataset}%
\end{table}%

Human demonstration data was collected to construct $\mathcal{D}_{\text{demo}}$, which was used to enhance the training of HGCR-DDPG. This chapter systematically evaluated the effectiveness of three key technical improvements (HTSK, GABC, CPL) introduced in the HGCR-DDPG algorithm and their contributions to model performance through comparative experiments with multiple algorithms. These comparative algorithms include:

1. \textbf{Rainbow-DDPG}: As the baseline model for the experiment, Rainbow-DDPG utilizes the same neural network structure to implement $\mu(\boldsymbol{s}_t,p_{\text{g}_t};\boldsymbol{\theta}^{\mu})$, $Q(\boldsymbol{s}_t,\boldsymbol{a}_t;\boldsymbol{\theta}^q)$, and the target networks $\mu^\prime$, $Q^\prime_1$ and $Q^\prime_2$. These networks consist of an input layer, 50 hidden fully connected layers (each with 16 neurons), and an output layer. In this structure, the loss function $\mu(\boldsymbol{s}_t,p_{\text{g}_t};\boldsymbol{\theta}^{\mu})$ is defined as (\ref{eq:used_lbc}).

The original Rainbow-DDPG does not use the HTSK fuzzy system but employs a neural network as $H(\boldsymbol{s}_t;\boldsymbol{\theta}^h)$. 
During the training process, a cross-entropy loss function is used as the BC loss function $L_{\text{bc}}$ for $H(\boldsymbol{s}_t;\boldsymbol{\theta}^h)$, defined as follows:

\begin{equation}
	L_{\text{bc}}=\begin{cases}
		-\sum_{p=1}^k\delta(p,p_{\text{g}_i})\log(H_p), & Q(\boldsymbol{s}_i,\boldsymbol{a}_i) \in \mathcal{D}_{\text{grasp}} \\
		0, & \text{otherwise}
	\end{cases},        \label{eq:cross_entropy_bcloss}
\end{equation}
where $k$ is the number of candidate endpoints. $\delta(p,p_{\text{g}_i})$ is the Kronecker delta function, which equals 1 when $p=p_{\text{g}_i}$, otherwise it's 0. $H_p$ is the probability corresponding to the $p$-th endpoint in the output probability distribution $H(\boldsymbol{s}_i;\boldsymbol{\theta}^h)$.It's assumed that the grasp point selected from $\mathcal{D}_{\text{grasp}}$ is consistently superior to $H(\boldsymbol{s}_t;\boldsymbol{\theta}^h)$. The cross-entropy loss function is applied whenever the sampled data $(\boldsymbol{s}_i,\boldsymbol{a}_i)$ comes from $\mathcal{D}_{\text{grasp}}$; otherwise, $L_{\text{bc}}$ is set to 0. Note that in Rainbow-DDPG, the training strategy does not include CPL. This means $H(\boldsymbol{s}_t;\boldsymbol{\theta}^h)$ is trained in the same way as the current Actor, and its output is not used as part of the input for $\mu(\boldsymbol{s}_t,p_{\text{g}_t};\boldsymbol{\theta}^{\mu})$.

2. \textbf{Rainbow-DDPG + GABC}: Integrates only the GABC into Rainbow-DDPG.

3. \textbf{Rainbow-DDPG + CPL + HTSK}: Incorporates the CPL into Rainbow-DDPG, using HTSK to select the grasp point.

4. \textbf{Rainbow-DDPG + CPL + Random}: Integrates the CPL into Rainbow-DDPG, employing a random sampling strategy to select the grasp point.

5. \textbf{Rainbow-DDPG + CPL + Uniform}: Combines the CPL with Rainbow-DDPG, utilizing a uniform sampling strategy to select the grasp point.

6. \textbf{Rainbow-DDPG + GABC + CPL + Random}: Merges both GABC and CPL into Rainbow-DDPG, employing a random sampling strategy to select the grasp point.

7. \textbf{Rainbow-DDPG + GABC + CPL + Uniform}: Extends Rainbow-DDPG by incorporating both GABC and CPL, selecting the grasp point using a uniform sampling strategy.

To evaluate the performance under different levels of difficulty, we designed two modes, simple and challenging, for each of the three tasks. In the simple mode, the maximum number of operations for each task is set to $t_\text{m}$; while in the challenging mode, this limit is halved to $\frac{t_\text{m}}{2}$ ($t_\text{m}$ is 2 for fold along diagonal tasks, 4 for fold along axis tasks, and 10 for flatten tasks). Additionally, to investigate the specific impact of the amount of human demonstration data, 
the models were trained with human demonstration data from 5, 20, and 100 rounds, respectively.

During the training phase, we first performed 20 rounds of pre-training to optimize the Critic network using GABC technique. Subsequently, training proceeded to the regular phase, which consisted of 30 training epochs, with each epoch comprising 20 rounds of training and a batch size ($N$) of 64. After executing a set of actions (grasping and manipulation), the policy was updated $t_{\text{n}}$ times. The single interaction update counts ($N$) for policies in fold along diagonal tasks, fold along axis tasks, and flatten tasks were 80, 40, and 20, respectively.

At the end of each training epoch, we conducted a testing phase comprising 10 rounds. Specifically, this study initially tested the performance of the initial policy upon completing policy initialization and increased the testing frequency during the pre-training phase (testing every 5 rounds of training). 
Ultimately, we conducted 35 testing epochs. Furthermore, 
we conducted experiments with three different random seeds. 
At the end of the $t$-th testing epoch ($t=0,\ldots,34$), for each seed $i$ ($i=1,2,3$), we recorded the total reward $R_j^{i,t},j=1,\ldots,10$ obtained by the agent in a single round. Subsequently, the average reward $R_{\text{avg}}^{i,t}$ for each seed was computed across the ten testing instances.

Based on the aforementioned processing steps, we define several key performance metrics:

1. The average reward per testing epoch $R_{\text{avg}}^t$ is defined as the average of all $R_{\text{avg}}^{i,t}$ values within the $t$-th testing epoch. All reward curves presented in this article are plotted based on $R_{\text{avg}}^{i,t}$. The specific calculation formula is:

\begin{equation}
	R_{\text{avg}}^t = \frac{1}{3}\sum_{i=1}^{3} R_{\text{avg}}^{i,t}.
\end{equation}

2. The average reward $R_{\text{avg}}$, which is the average of all $R_t^{avg}$ values for $t = 0, \ldots, 34$.

3. The average standard deviation $\sigma_{\text{avg}}$. Firstly, calculate the standard deviation $\sigma^t$ for $R_{\text{avg}}^{i,t}$ for $i = 1, 2, 3$, then calculate the average of all $\sigma^t$ as $\sigma_{\text{avg}}$. 

4. Average Reward Ranking $Rank_{\text{avg}}$ can be obtained by ranking all algorithms based on $R_{\text{avg}}$.

5. Average Standard Deviation Ranking $Rank_{\sigma}$ can be obtained by ranking all algorithms based on $\sigma_{\text{avg}}$. Lower rankings demonstrate higher stability. 


\subsection{Experiment Settings for Verifying the Effectiveness of the NMPC Demonstration Dataset}

We constructed a $6\times6$ point-mass spring-damper model, collected demonstration data, and selected the top 100 rounds with the highest final rewards for each task, forming $\mathcal{D}_{\text{NMPC}}$. Table \ref{tab:NMPC_performance} provides a detailed analysis of the NMPC demonstration dataset for three specific tasks. 
Compared to Table \ref{tab:Human Demonstration Dataset}, the NMPC dataset exhibits similar average rewards and standard deviations but need more average number of steps to complete tasks, indicating that NMPC strategies can accomplish tasks with lower operational efficiency.

\begin{table}[tb]
	\centering
	\caption{Human Demonstration Dataset}
	\begin{tabular}{c|ccc}
		\toprule
		\diagbox [width=6em,trim=l] {Task}{Metric} & \begin{tabular}[c]{@{}c@{}}
			Average\\Reward
		\end{tabular} & \begin{tabular}[c]{@{}c@{}}
			Reward\\Standard Deviation
		\end{tabular} & \begin{tabular}[c]{@{}c@{}}
			Average\\Steps
		\end{tabular} \\
		\hline
		\begin{tabular}[c]{@{}c@{}}
			Folding Along\\the Diagonal
		\end{tabular} & 95.1 & 2.4 & 4.4 \\
		\begin{tabular}[c]{@{}c@{}}
			Folding Along\\the Central Axis
		\end{tabular}& 87.1 & 3.5 & 6.1 \\
		Flattening& 80.9 & 6.7 & 9.2 \\
		\bottomrule
		\hline
	\end{tabular}%
	\label{tab:NMPC_performance}%
\end{table}%


In this experiments, we adopt the same experimental settings, evaluation metrics, and numbering system as in Section IV.B. The demonstration dataset $\mathcal{D}_{\text{NMPC}}$ generated by the NMPC algorithm is used as the demonstration training set for the HGCR-DDPG model. Three different statistical metrics are also employed in this experiment to compare the effectiveness of HGCR-DDPG models trained with assistance from $\mathcal{D}_{\text{NMPC}}$ and $\mathcal{D}_{\text{demo}}$. These metrics are Cosine Similarity (CSS), Dynamic Time Warping (DTW), and Pearson Correlation Coefficient (PCC). We have calculated these three similarity measures in terms of both reward and standard deviation, the specific metrics are as follows:

1. Reward Cosine Similarity (RCS): Cosine similarity of the average reward sequence $R_{\text{avg}}^t$ for a single test cycle.

2. Reward Dynamic Time Warping (RDT): DTW similarity of the average reward sequence $R_{\text{avg}}^t$ for a single test cycle.

3. Reward Pearson Correlation (RPC): Pearson correlation coefficient of the average reward sequence $R_{\text{avg}}^t$ for a single test cycle.

4. Standard Deviation Cosine Similarity (SCS): Cosine similarity of the standard deviation of reward sequences obtained with different random seeds.

5. Standard Deviation Dynamic Time Warping (SDT): DTW similarity of the standard deviation of reward sequences obtained with different random seeds.

6. Standard Deviation Pearson Correlation (SPC): Pearson correlation coefficient of the standard deviation of reward sequences obtained with different random seeds.

\subsection{Physical Experiment for Deformable Object Robot Manipulation}

Hand-eye calibration is used to determine the spatial relationship between the camera coordinate system and the robot coordinate system. Endpoint detection is utilized to extract key endpoints from the fabric's image. Optical flow tracking is employed for real-time tracking of the fabric endpoints' positions. The specific processes of these three parts are as follows:

\subsubsection{Hand-eye calibration}
In this study, the easy-eye-hand software package is utilized for calibration.

\subsubsection{Endpoint Detection}
During the initial stages of the tasks, this study employs endpoint detection algorithms to extract the pixel coordinates of fabric endpoints in the image. 
For non-initial states in folding tasks, we use the previous placement point $\boldsymbol{p}_{\text{p}_{t-1}}$ as a reference to determine the current position of the baseline grasp point. Simultaneously, optical flow tracking algorithms are employed to track the movement of the remaining endpoints. 
After coordinate transformation and hand-eye calibration, the three-dimensional positions of the endpoints in the robot coordinate system are obtained. Combining observable visual information such as the fabric's area, a comprehensive state vector is formed.


We designed an endpoint recognition algorithm for extracting key endpoints from images of fabrics. The algorithm is based on the Canny edge detection and Douglas-Peucker polygon approximation algorithms, which accurately extract edge information from fabrics and identify several furthest endpoints. First, the input RGB image is converted into a grayscale image. Then, Gaussian blur is applied to the grayscale image to reduce noise influence. Next, the Canny edge detection algorithm is employed to extract the edges of the image. We further refine the edges by contour detection to find the largest contour in the image and apply the Douglas-Peucker algorithm \cite{c32} for polygon approximation to simplify the contour. Finally, we obtain a simplified contour containing a series of points, denoted as $\boldsymbol{C} = (\boldsymbol{p}_1, \boldsymbol{p}_2, \ldots, \boldsymbol{p}_n)$.

To select the furthest $k$ endpoints from the simplified contour $\boldsymbol{C}$ (where $k=4$ for folding tasks and $k=8$ for flattening tasks), this study devises a heuristic method called Maximum Minimum Distance Vertices Selection (MMDVS). 
For any $k$ points $\boldsymbol{p}_{r_1}, \boldsymbol{p}_{r_2}, \cdots, \boldsymbol{p}_{r_k}$ on the contour, we first define a set $\boldsymbol{S}$ containing all possible pairs of points. Then, we compute the Euclidean distance between each pair of points in set $\boldsymbol{S}$ and find the minimum distance $d_{min}$. By traversing all possible combinations of $k$ points in $\boldsymbol{C}$, we find the group of points with the maximum minimum distance $d_{min}$, which are the desired $k$ endpoints $\boldsymbol{P} = (\boldsymbol{p}_1, \boldsymbol{p}_2, \cdots, \boldsymbol{p}_k)$. Additionally, if the number of endpoints in the simplified contour $\boldsymbol{C}$ obtained by the Douglas-Peucker algorithm is less than $k$, we adjust the relevant parameters of the Douglas-Peucker algorithm to include new endpoints into the simplified contour set $\boldsymbol{C}$ until a sufficient number of endpoints are obtained. 
The area of the fabric, needed for deep RL, can be obtained by calculating the area of the contour, and the centroid can be obtained by calculating the centroid of the contour by using relevant functions in the OpenCV library. 

\subsubsection{Optical Flow Tracking}
To achieve real-time tracking of the fabric's shape, this study designs an optical flow-based tracking algorithm. To address the issue that tracking failures can be attributed to occlusion by the end effector, 
the endpoints that cannot be successfully tracked can be classified into two categories: non-baseline grasp points and baseline grasp points. For non-baseline grasp points, their positions are set to the positions from the last frame before the tracking failure, $\boldsymbol{p}_{i_{t-1}}, i=1,2,\cdots,k$, and tracking continues based on these positions. For baseline grasp points, after completing the placement action, the placement point $\boldsymbol{p}_{\text{p}_{t-1}}$ is considered as the new position for anchor grasp points.

The comprehensive experimental procedure is shown in Fig. S5. At the beginning of the experiment, the robot and fabric were placed in their initial states. Subsequently, the system operated in a loop according to the following steps until the fabric is manipulated from its initial state to the target state. First, capture and analyze the current state of the fabric to form a comprehensive state vector. Then, this state vector is input into the HGCR-DDPG model pre-trained with $\mathcal{D}_{\text{NMPC}}$ to generate instructions. Next, the robot executes actions according to the generated instructions, pushing the fabric towards the desired next state. Finally, after the operation is completed, the system collects and updates the state vector of the fabric, preparing for the next steps of operation.


\section{Experiment Results}

\subsection{Simulation Results of the Improved HGCR-DDPG Algorithm with Human Demonstrations}
\begin{figure*}[htbp]
	\centering
	\includegraphics[width=0.9\textwidth]{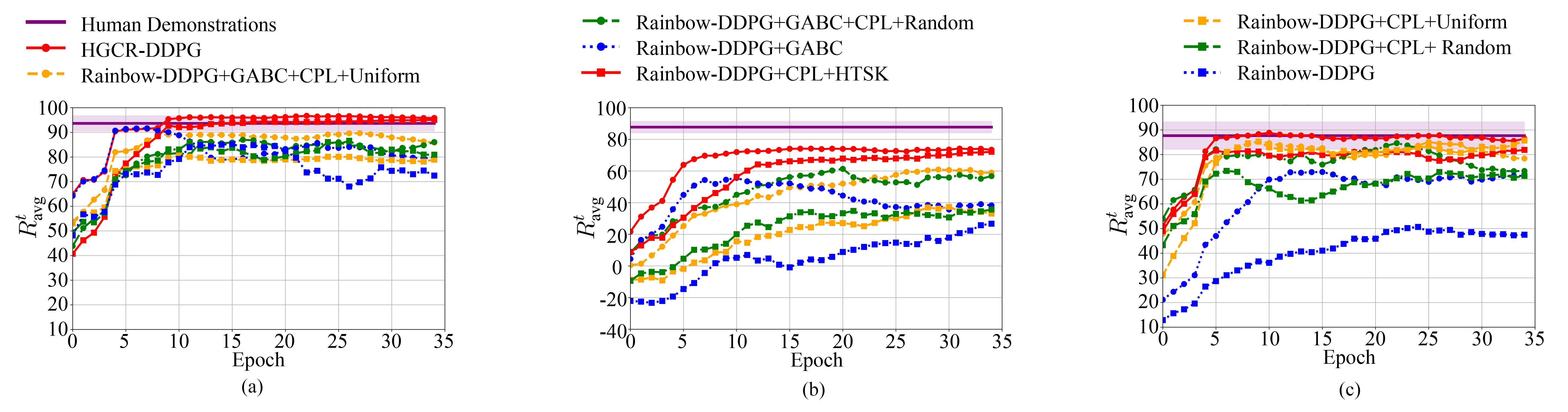}
	\caption{Reward Curve. (a) Folding along the diagonal. (b) Folding along the central axis. (c) Flattening.}
	\label{Reward Curve}
\end{figure*}

Tables S1 and S2 respectively list the algorithm numbers and experiment numbers involved in this article. Table S3 presents the design of 8 control groups in this study. Each control group evaluates the effectiveness of three technical improvements introduced in the HGCR-DDPG algorithm (HTSK, GABC, CPL) by comparing the performance of multiple algorithms. Table S3 also lists the legend styles and expected results of each algorithm in Fig. \ref{Reward Curve} and Fig. S6.

Fig. \ref{Reward Curve} and Fig. S6 display the dynamic changes of the average reward $R_{\text{avg}}^t$ of individual test cycles obtained by various algorithms in different simulation experiments under three task settings: folding along the diagonal, folding along the central axis, and flattening. 
We smoothed the reward curves using a window of length 3. Tables S4, S6, and S8 respectively list the average rewards $R_{\text{avg}}$ and average reward rankings $Rank_{\text{avg}}$ achieved by various algorithms in different experiments conducted under the three tasks. Tables S2, S7, and S9 then respectively display the average standard deviations $\sigma_{\text{avg}}$ and average standard deviation rankings $Rank_{\sigma}$ achieved by various algorithms in different experiments targeting the aforementioned three tasks. The last column of each of these tables shows the average value of the respective metrics ($R_{\text{avg}}/Rank_{\text{avg}}$ or $\sigma_{\text{avg}}/Rank_{\sigma}$) for each algorithm across all experiments for that task. Table \ref{tab:algorithm_rankings} presents the global average reward (average of $R_{\text{avg}}$), global average standard deviation ($\sigma_{\text{avg}}$), reward average ranking (average of $Rank_{\text{avg}}$), and standard deviation average ranking (average of $Rank_{\sigma}$) for each algorithm across all experiments. In this section, the optimal indicators are highlighted in bold, and the HGCR-DDPG algorithm proposed in this article is underlined.

\begin{table}[tb]
	\centering
	\caption{Global Performance Metrics and Rankings of Algorithms}
	\begin{tabular}{ccccc}
		\toprule
		\begin{tabular}[c]{@{}c@{}}
			Algorithm\\ Code
		\end{tabular} & \begin{tabular}[c]{@{}c@{}}
			Global \\ Average\\Reward
		\end{tabular} & \begin{tabular}[c]{@{}c@{}}
			Global \\Average\\Standard\\ Deviation
		\end{tabular} & \begin{tabular}[c]{@{}c@{}}
			Average\\Reward\\ Rank
		\end{tabular} & \begin{tabular}[c]{@{}c@{}}
			Average \\Standard\\Deviation\\ Rank
		\end{tabular} \\
		\hline
		\underline{1} & \textbf{76.3} & \textbf{4.8} & \textbf{1.1} & \textbf{7.1}\\
		2 & 60.1 & 11.2 & 3.6 & 3.9 \\
		3 & 57.1 & 11.9 & 4.8 & 3.3 \\
		4 & 57.8 & 8.1 & 4.7 & 5.2 \\
		5 & 67.4 & 6.7 & 2.7 & 6.4 \\
		6 & 52.2 & 14.0 & 5.7 & 2.7 \\
		7 & 52.4 & 12.7 & 5.9 & 2.7 \\
		8 & 37.9 & 10.6 & 7.5 & 4.8\\
		\bottomrule
	\end{tabular}%
	\label{tab:algorithm_rankings}%
\end{table}%

From these figures and tables, we can derive the following key conclusions:
Firstly, Algorithm 1 (HGCR-DDPG) demonstrates outstanding performance across all numerical simulation experiments for all tasks.
Secondly, HTSK significantly enhances algorithm performance. Under the same marker style, the red curve (representing algorithms using HTSK) generally exhibits significant advantages. In the majority of experimental scenarios, algorithms utilizing HTSK for benchmark grasping point selection outperform those employing random selection strategies, particularly in experimental setups with stringent constraints on the number of operations.
Furthermore, the positive impact of GABC is also significant. Among curves of the same color, the curves marked with circles (representing algorithms using the GABC) generally outperform those marked with squares. Although the performance of algorithms incorporating GABC in flattening tasks is not particularly remarkable in terms of standard deviation, its performance surpasses algorithms without GABC in all other experiments.
Additionally, the effectiveness of CPL has been validated. It is worth noting that the performance of CPL is influenced by both the benchmark grasping point selection strategy it adopts and the operational constraints in the experiments. The looser the constraints on the number of operations and the more stable the benchmark grasping point selection strategy, the more significant the effect of CPL.
Finally, from the Table \ref{tab:algorithm_rankings}, it can be observed that Algorithm 1 (HGCR-DDPG) achieved the best performance across all metrics. Compared to the selected baseline algorithm, namely Algorithm 8 (Rainbow-DDPG), HGCR-DDPG achieved a 2.01-fold improvement in global average reward and successfully reduced the global average standard deviation to 45\% of the baseline algorithm, demonstrating a significant performance advantage.

\subsection{Results of the Experiment for Verifying the Effectiveness of the NMPC Demonstration Dataset}

\begin{figure*}[htbp]
	\centering
	\includegraphics[width=0.9\textwidth]{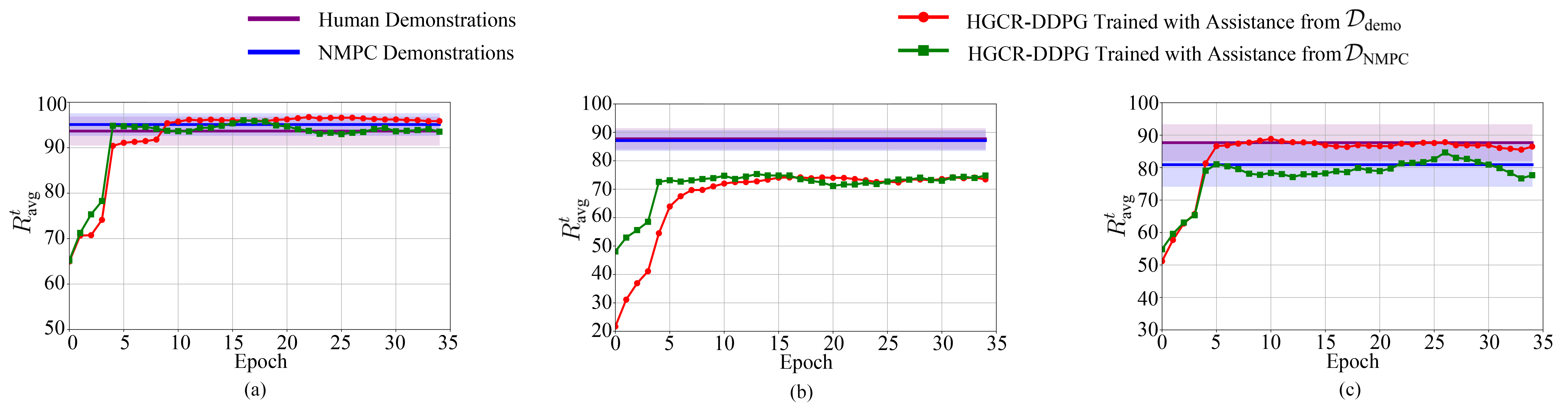}
	\caption{Performance Comparison of HGCR-DDPG Trained with Assistance from $\mathcal{D}_{\text{NMPC}}$ and $\mathcal{D}_{\text{demo}}$. (a) Folding along the diagonal. (b) Folding along the central axis. (c) Flattening.}
	\label{Performance Comparison}
\end{figure*}

Fig. \ref{Performance Comparison} and Fig. S7 depict the variation curves of the average reward $R_{\text{avg}}^t$ for single test cycles of the HGCR-DDPG model trained with assistance from $\mathcal{D}_{\text{NMPC}}$ and $\mathcal{D}_{\text{demo}}$ for the tasks of folding along the diagonal, folding along the central axis, and flattening. Tables S10, S11, and S12 respectively show the performance of the HGCR-DDPG model assisted by $\mathcal{D}_{\text{NMPC}}$ in terms of average reward $R_{\text{avg}}$ and average standard deviation $\sigma_{\text{avg}}$ for the three tasks, as well as the ratio of the performance achieved by models assisted by $\mathcal{D}_{\text{NMPC}}$ to those assisted by $\mathcal{D}_{\text{demo}}$.

From these curves and tables, it is evident that in the task of folding along the diagonal, HGCR-DDPG can quickly learn and develop effective strategies regardless of whether $\mathcal{D}_{\text{NMPC}}$ or $\mathcal{D}_{\text{demo}}$ is used. However, in the tasks of folding along the central axis and flattening, as the difficulty increases, the performance difference between HGCR-DDPG assisted by the two demonstration datasets gradually becomes significant. In simplified task settings (Experiments 2.2, 2.4, 2.6, 3.2, 3.4), HGCR-DDPG assisted by $\mathcal{D}_{\text{NMPC}}$ demonstrates the ability to learn rapidly, with its performance even reaching or slightly exceeding that of models assisted by $\mathcal{D}_{\text{demo}}$. This may be because the NMPC strategy itself performs well in scenarios with a generous number of steps, allowing HGCR-DDPG to effectively extract strategies from its demonstrations. Conversely, under more stringent task settings (Experiments 2.1, 2.3, 2.5, 3.3, 3.5), HGCR-DDPG assisted by $\mathcal{D}_{\text{NMPC}}$ is generally lower than models assisted by $\mathcal{D}_{\text{demo}}$. This could be attributed to the difficulty of the NMPC strategy in completing tasks within a limited number of steps, thus affecting the performance of HGCR-DDPG under these conditions. This pattern also aligns with the higher average number of steps observed in the NMPC dataset in Table \ref{tab:NMPC_performance}. The results of Experiments 3.1 and 3.6 show the inherent stochastic factors in the experimental process.

\begin{table}[tb]
	\centering
	\caption{Comparison of Global Performance Metrics for HGCR-DDPG Trained with $\mathcal{D}_{\text{NMPC}}$ and $\mathcal{D}_{\text{demo}}$}
	\begin{tabular}{c|ccc}
		\toprule
		\begin{tabular}[c]{@{}c@{}}
			Demonstration\\Dataset
		\end{tabular} & \begin{tabular}[c]{@{}c@{}}
			Global Average\\Reward
		\end{tabular} & \begin{tabular}[c]{@{}c@{}}
			Global Average\\Standard Deviation
		\end{tabular}\\
		\hline
		$\mathcal{D}_{\text{demo}}$ & 76.3 & 4.8\\
		$\mathcal{D}_{\text{NMPC}}$ & 76.1 & 4.0\\
		\bottomrule
	\end{tabular}%
	\label{tab:global_performance_comparison}%
\end{table}%

Table \ref{tab:global_performance_comparison} presents a comparison of the overall performance metrics between HGCR-DDPG models assisted by $\mathcal{D}_{\text{NMPC}}$ and $\mathcal{D}_{\text{demo}}$. The global average reward achieved by the HGCR-DDPG model assisted by $\mathcal{D}_{\text{NMPC}}$ is 99.7\% of that achieved by the model assisted by $\mathcal{D}_{\text{demo}}$, while the global average standard deviation of rewards obtained under different random seeds is 83.3\% of that achieved by the model assisted by $\mathcal{D}_{\text{demo}}$. This indicates that HGCR-DDPG assisted by $\mathcal{D}_{\text{NMPC}}$ exhibits a performance level similar to that of HGCR-DDPG assisted by $\mathcal{D}_{\text{demo}}$.

From Table S13, it can be observed that the HGCR-DDPG model trained with $\mathcal{D}_{\text{NMPC}}$ and $\mathcal{D}_{\text{demo}}$ exhibits significant similarity at the $R_{\text{avg}}^t$ sequence level, particularly in RCS and RPC. This emphasizes a strong consistency between the $\mathcal{D}_{\text{NMPC}}$-assisted HGCR-DDPG model and the $\mathcal{D}_{\text{demo}}$-assisted model concerning the $R_{\text{avg}}^t$ sequence. However, in terms of the standard deviation of reward sequences obtained with different random seeds, the similarity metrics show significant differences, especially evident in SPC. This difference may stem from two factors: firstly, the inherent randomness of the experiment may lead to some fluctuations in reward curves under different random seeds; secondly, the stylistic differences between NMPC and human-operated strategies may cause the model to adopt different action strategies in specific contexts, thus affecting certain performance metrics.

\subsection{Results of Physical Experiments on Visual Processing and Robot Manipulation}

\begin{figure}[tb]
	\centering
	\includegraphics[width=0.5\textwidth]{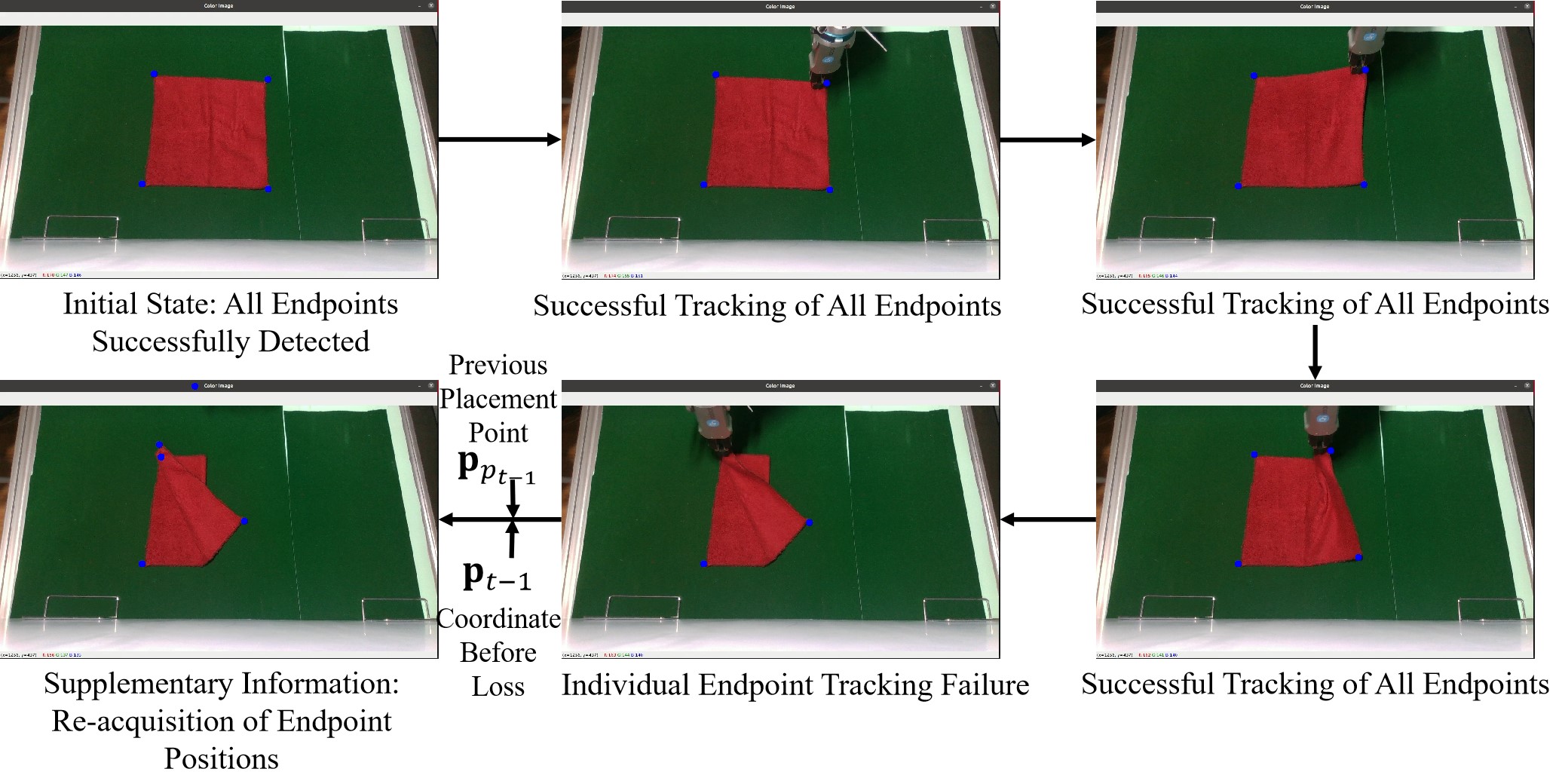}
	\caption{Optical Flow Tracking Results.}
	\label{Optical Flow Tracking Results}
\end{figure}

The endpoint recognition algorithm mainly targets two situations: when the fabric is completely flattened and when it is fully wrinkled. In the case of complete flattening, the endpoint recognition algorithm can accurately identify the four endpoints of the fabric, as shown in the first picture of Fig. \ref{Optical Flow Tracking Results}. In the case of complete wrinkling, the endpoint recognition algorithm can also accurately identify the eight representative endpoints of the fabric, as shown in Fig. S8. This indicates that the endpoint recognition algorithm can accurately identify the endpoints of the fabric under different fabric states, providing accurate initial positions for subsequent optical flow tracking.
The optical flow tracking algorithm is primarily used to track the endpoints of the fabric in folding tasks, as shown in Fig. \ref{Optical Flow Tracking Results}. This indicates that the optical flow tracking algorithm can accurately track the endpoints of the fabric when the shape of the fabric changes, providing precise target positions for subsequent robot operations.
It is worth noting that in the second-to-last image of Fig. \ref{Optical Flow Tracking Results}, there is a tracking failure for the two endpoints in the top left corner of the fabric. This is caused by occlusion from the end effector, and in such cases, we utilized the method introduced in Section IV.D for supplementation. After the operation is completed, the placement point is treated as the new position of the reference grasping point, ensuring the smoothness of robot operations.

The experimental operation process is illustrated in Fig. S9. In physical experiments, measuring the distance between different endpoints is inconvenient, so several indicators directly computable from visual information were set as follows:

1. Task Completion Rate: For folding along the diagonal, the target shape of the fabric was set as an isosceles right triangle with a side length of 0.24 meters. For folding along the central axis, the target shape of the fabric was set as a rectangle measuring 0.24 meters by 0.12 meters. For flattening, the target shape of the fabric was set as a square with a side length of 0.24 meters. Subsequently, the similarity between the target shape of the fabric and the actual shape was calculated using the `cv2.matchShapes()' function in OpenCV, and this was used as the task completion rate.

2. Success Rate: An experiment was considered successful when the final task completion rate exceeded 0.9.

3. Average Steps: The average number of actions required for the robot to complete a specific task measured the efficiency of the robot's operations. In this study, due to the thinness of the fabric used, the positioning accuracy of the sensor subsystem in the z-axis direction was extremely strict, with a tolerance of only 2 mm, which greatly increased the likelihood of gripping failure. To address this challenge, a heuristic strategy was adopted in the experiment: first attempt gripping based on the positioning information provided by the sensor subsystem. If the first gripping attempt was unsuccessful (i.e., no improvement in task completion rate), the gripping point was lowered by 2 mm in the z-axis direction and another attempt was made, repeating this process until successful gripping was achieved.

In the process of counting operation steps, this study only included each successful placement action in the total steps, without counting repeated attempts due to gripping failures. 
30 experiments were conducted for each of the three tasks, and the aforementioned indicators were recorded. The experimental results are shown in Table \ref{tab:physical_experiment_results}. For the folding along the diagonal task, 93.3\% of the trials achieved a task completion score of no less than 0.6, while 90.0\% of the trials achieved a task completion score of no less than 0.8. The overall success rate for this task is 83.3\%, with an average of 1.1 steps required, indicating a relatively high success rate and fewer required steps. For the folding along the central axis task, 90.0\% of the trials reached the standard of a task completion score of at least 0.6, and 86.7\% of the trials reached the standard of a task completion score of at least 0.8. The success rate is 80.0\%, with an average of 3.9 steps required. Compared to folding along the diagonal, this task requires more steps but still maintains a relatively high success rate. For the flattening task, all trials reached the standards of a task completion score of at least 0.6 and 0.8, with a high success rate of 96.7\%. However, the average number of steps required is 13.5 steps, indicating that although the flattening task has the highest success rate, it is also the most time-consuming of the three tasks. This phenomenon can be attributed to the high tolerance for errors in the flattening task. Specifically, even if a certain operation leads to a decrease in task completion score, the robot can still flatten the fabric through subsequent operations. This characteristic leads to a high success rate for the flattening task but also results in an increase in the number of required steps. In summary, the success rates of all three tasks are relatively high, indicating that the experimental setup and methods used perform well in physical operations.

\begin{table*}[tb]
	\centering
	\caption{Physical Experiment Results}
	\begin{tabular}{c|cccc}
		\toprule
		\diagbox [width=6em,trim=l] {Task}{Metric}  & Task Completion $\geq 0.6$ & Task Completion $\geq 0.8$ & Success Rate & Average Steps \\
		\hline
		Diagonal Folding & 93.3\% & 90.0\% & 83.3\% & 1.1\\
		Central Axis Folding & 90.0\% & 86.7\% & 80.0\% & 3.9\\
		Flattening & 100.0\% & 100.0\% & 96.7\% & 13.5\\
		\bottomrule
	\end{tabular}%
	\label{tab:physical_experiment_results}%
\end{table*}%

\subsection{Discussion}

In the context of robotic manipulation tasks for deformable objects, this article addresses the inefficiency of traditional RL methods by proposing the HGCR-DDPG algorithm. To tackle the issue of high costs associated with traditional human teaching methods, a low-cost demonstration collection method based on NMPC is introduced. The effectiveness of the proposed methods is validated through three experimental scenarios involving folding fabric diagonally, along the midline, and flattening it, both in simulation and real-world experiments. Extensive ablation studies are conducted to substantiate the rationality and efficacy of the algorithms.

Compared to similar research, Matas et al. \cite{c18} required nearly 80,000 interactions between the robot and the environment to complete the learning process; Jangir et al. \cite{c19} needed approximately 260,000 rounds of interaction data to train their agent; Yang et al. \cite{c24} utilized 28,000 pairs of images and actions collected via teleoperation to train a DNN as an end-to-end policy for folding a single towel. This study simplifies the data acquisition process and achieves comparable or even higher success rates than the aforementioned studies, providing novel insights and contributions for future tasks of a similar nature. Currently, more and more research tends to adopt Vision-Language-Action models (VLA) for robotic manipulation. However, such research often requires significant computational resources and is overqualified when dealing with specific tasks. For example, OpenVLA is a 7B-parameter VLA that was trained on 64 A100 GPUs for 14 days. During inference, it requires 15GB of video memory and runs at approximately 6Hz on an NVIDIA RTX 4090 GPU \cite{c33}. The largest model of RT-2 uses 55B parameters, and it is infeasible to directly run such a model on standard desktop machines or on-robot GPUs commonly used for real-time robot control \cite{c34}. Even TinyVLA requires 1.3B parameters \cite{c35}. In contrast, our proposed algorithm shows significant advantages in learning efficiency. Trained on an Intel i5 12400f CPU and NVIDIA RTX 3050 GPU, our algorithm can converge within dozens of epochs, and the entire training process takes at about 4 hours, with a maximum number of 14,183 parameters. Compared with the currently popular approaches based on large models for robot manipulation, the algorithm proposed in this paper has the advantages of being lightweight, requiring low computational resources, and being able to provide task-specific customization and efficient adaptability when handling specific tasks.

\section{Conclusion}
This article presents a study on deformable object robot manipulation based on demonstration-enhanced RL. 
To improve the learning efficiency of RL, this article enhances the utilization efficiency of algorithms for demonstration data from multiple aspects, proposing the HGCR-DDPG algorithm and collecting $\mathcal{D}_{\text{demo}}$ for training. It first uses demonstration data to train the HTSK fuzzy system to select appropriate grasp points, then proposes the GABC to improve the utilization of demonstration data in Rainbow-DDPG, and finally uses CPL to synthesize HTSK and GABC improved Rainbow-DDPG, forming a complete control algorithm for deformable object robot manipulation, namely HGCR-DDPG. Additionally, the effectiveness of the proposed methods is verified through comprehensive simulation experiments. Compared to the baseline algorithm (Rainbow-DDPG), the proposed HGCR-DDPG algorithm achieves a 2.01 times higher global average reward and reduces the global average standard deviation to 45\% of the baseline algorithm. To reduce the labor cost of demonstration collection, this article proposes a low-cost demonstration collection method based on NMPC. Based on the established spring-mass model, it uses the NMPC algorithm to control the robot to perform deformable object manipulation tasks in a simulation environment, and uses the trajectories of rounds with higher rewards as demonstration data. Simulation results show that the global average reward obtained by the HGCR-DDPG model trained with $\mathcal{D}_{\text{NMPC}}$ is 99.7\% of the model trained with $\mathcal{D}_{\text{demo}}$, and the global average standard deviation of rewards obtained under different random seeds is 83.3\% of the model trained with $\mathcal{D}_{\text{demo}}$. This indicates that demonstration data collected through NMPC can be used to train HGCR-DDPG and its effectiveness is comparable to human demonstration data. To verify the feasibility of the proposed methods in a real environment, this article conducts physical experiments on deformable object robot manipulation. Utilizing hardware facilities such as the UR5e robot, OnRobot RG2 gripper, and RealSense D435i camera, this article builds a physical experimental platform for deformable object robot manipulation and uses the $\mathcal{D}_{\text{NMPC}}$-assisted training HGCR-DDPG algorithm on this platform to control the robot to manipulate fabric and perform folding along the diagonal, folding along the central axis, and flattening tasks. The experimental results show that the proposed methods achieve success rates of 83.3\%, 80\%, and 100\% respectively in these three tasks, verifying the effectiveness of the method.

There are still many areas for improvement due to time constraints. Specifically, future work of this article could be expanded in the aspect such as multimodal perception input for RL state vectors, refinement of deformable object dynamic models, and small-sample learning for operations on various deformable objects, etc.

\vfill

\end{document}